\documentclass[pdflatex,sn-mathphys-num]{sn-jnl}

\usepackage{rotating}
\usepackage{array}
\usepackage{float}

\usepackage{graphicx}%
\usepackage{multirow}%
\usepackage{amsmath,amssymb,amsfonts}%
\usepackage{amsthm}%
\usepackage{mathrsfs}%
\usepackage[title]{appendix}%
\usepackage{xcolor}%
\usepackage{textcomp}%
\usepackage{manyfoot}%
\usepackage{booktabs}%
\usepackage{algorithm}%
\usepackage{algorithmicx}%
\usepackage{algpseudocode}%
\usepackage{listings}%
\usepackage{comment}%
\usepackage{hyperref}

\usepackage{lscape}

\theoremstyle{thmstyleone}%
\theoremstyle{thmstyletwo}%

\theoremstyle{thmstylethree}%

\begin{document}

\title{Towards General Urban Monitoring with Vision-Language Models: A Review, Evaluation, and a Research Agenda} 

\author*[1,2]{\fnm{André} \sur{Torneiro}}\email{andre.torneiro12@gmail.com}

\author[1]{\fnm{J. Diogo} \sur{Monteiro}}\email{jdiogoxmonteiro@gmail.com}

\author[1]{\fnm{Paulo} \sur{Novais}}\email{pjon@di.uminho.pt}

\author[1]{\fnm{Pedro} \sur{Rangel Henriques}}\email{pjon@di.uminho.pt}

\author[3,4]{\fnm{Nuno F.} \sur{Rodrigues}}\email{nfr@ipca.pt}

\affil*[1]{\orgdiv{ALGORITMI Research Centre/LASI}, \orgname{University of Minho}, \city{Braga}, \country{Portugal}}

\affil[2]{\orgname{Logimade}, \city{Funchal}, \state{Braga}, \country{Portugal}}

\affil[3]{\orgname{INESC TEC}, \city{Porto}, \country{Portugal}}

\affil[4]{\orgdiv{2Ai, School of Technology}, \orgname{IPCA}, \city{Barcelos}, \country{Portugal}}

\abstract{
Urban monitoring of public infrastructure (such as waste bins, road signs, vegetation, sidewalks, and construction sites) poses significant challenges due to the diversity of objects, environments, and contextual conditions involved. Current state-of-the-art approaches typically rely on a combination of IoT sensors and manual inspections, which are costly, difficult to scale, and often misaligned with citizens' perception formed through direct visual observation. This raises a critical question: Can machines now “see” like citizens and infer informed opinions about the condition of urban infrastructure? Vision-Language Models (VLMs), which integrate visual understanding with natural language reasoning, have recently demonstrated impressive capabilities in processing complex visual information, turning them into a promising technology to address this challenge.

This systematic review investigates the role of VLMs in urban monitoring, with particular emphasis on zero-shot applications. Following the PRISMA methodology, we analyzed 32 peer-reviewed studies published between 2021 and 2025 to address four core research questions: (1) What urban monitoring tasks have been effectively addressed using VLMs? (2) Which VLM architectures and frameworks are most commonly used and demonstrate superior performance? (3) What datasets and resources support this emerging field? (4) How are VLM-based applications evaluated, and what performance levels have been reported?

To the best of our knowledge, this is the first comprehensive review specifically dedicated to the application of VLM's in urban environment monitoring. Based on the knowledge gathered, we introduce a taxonomy of seven application categories, enabling researchers to quickly grasp the landscape of existing approaches and identify the most suitable methods for their specific tasks. This is followed by an in-depth analysis of the 11 most prominent VLM architectures, in which we highlight their respective strengths and limitations. We further examine the most widely used datasets, evaluation metrics, and benchmarking methodologies employed in the field. The article concludes with a forward-looking research agenda that identifies critical gaps and outlines strategic directions for future investigation.
}


    

\maketitle

\section{Introduction}\label{sec1}

Urban environments are evolving rapidly, shaped by growing populations, increasing infrastructure demands, and heightened sustainability goals. According to the United Nations, 68\% of the world’s population is projected to live in cities by 2050, up from 55\% in 2018 \cite{population}. This growing urbanization places immense pressure on urban infrastructure, safety, sustainability, and public services.

A critical enabler of this transformation is urban perception systems: AI-driven technologies that monitor city spaces for events such as illegal waste dumping, damaged infrastructure, unattended objects, unauthorized vehicle parking, and real-time traffic anomalies. These tasks, which were once labor-intensive or reliant on citizen reports, can now be partially automated through cameras, mobile devices, and Internet of Things (IoT) sensors. However, traditional computer vision systems, trained on fixed labeled datasets for narrow tasks, struggle to generalize across the diverse visual, linguistic, and cultural contexts found in urban scenes. Manual annotation of every possible class, in every language and geography, is impractical at scale. Moreover, these models often lack the flexibility to adapt to emerging events or novel object categories not seen during training.

Recent breakthroughs in Vision-Language Models (VLMs) and zero-/few-shot learning offer a promising alternative. Models like CLIP \cite{clip}, BLIP-2 \cite{blip2}, and Grounding DINO \cite{grounding} learn to associate images and language in a shared embedding space, allowing them to identify new concepts or behaviors via natural language prompts. This makes them particularly attractive for urban monitoring, where detection targets are numerous, ambiguous, and constantly evolving.

For example, given a prompt like "overflowing trash bin on the sidewalk", a zero-shot model can detect such instances without needing dedicated training samples. Similarly, a few-shot model can learn to recognize unauthorized e-scooter parking or damaged public benches from just a handful of examples. These capabilities open new avenues for scalable, multilingual, and dynamic urban AI systems.

Although prior surveys have explored zero-/few-shot learning in specific domains such as remote sensing \cite{remotereview} and autonomous driving \cite{autonomousreview}, a comprehensive review focused on urban vision-language tasks remains absent. These existing works typically emphasize domain-specific challenges, like satellite image segmentation or driving scene understanding, but do not address the broader spectrum of urban perception, including public infrastructure monitoring, multimodal prompts, and socio-spatial diversity.

This systematic review investigates the emerging intersection of vision-language models and urban AI through three key research interests:
\begin{itemize}
    \item \textbf{Application Landscape} – What urban tasks are currently being addressed using zero-shot or few-shot learning approaches with VLMs? Which areas (e.g., traffic, planning, surveillance, navigation) are most actively explored, and where are the gaps?

    \item \textbf{Methodological Innovation} – What types of models, prompts, and learning paradigms are employed? How are models adapted (or not) to urban contexts, and what strategies exist for overcoming data scarcity and domain shift?

    \item \textbf{Dataset and Deployment Realism} – What datasets underpin these systems? Are the models benchmarked for cross-city generalization, temporal robustness, or real-world deployment feasibility?
\end{itemize}
Following this introduction, the paper proceeds with a detailed methodology outlining the article selection strategy and inclusion criteria. Section 3 presents the results through task-based categorization and metric synthesis, while Section 4 offers a critical discussion of cross-cutting limitations, dataset patterns, and emerging trends. Finally, we conclude with a roadmap for future research that emphasizes scalable, inclusive, and deployment-ready urban AI.

This review contributes to the field by bridging VLM research and practical urban computing, offering a structured overview that spans technical innovation, dataset critique, and deployment foresight. It lays the groundwork for more inclusive, efficient, and real-world-ready urban AI systems built on the foundation of few-shot and zero-shot multimodal learning.

\section{Material \& Methods}\label{sec2}

The methodology used in this systematic review is based on the PRISMA technique \cite{bib14} and consists of 5 stages, as described below. The search strategy was developed using a structured framework to identify relevant studies that explore Vision Language Models (VLMs) in urban contexts. We defined the following key elements:

\begin{itemize}
    \item \textbf{Review object:} Studies involving VLMs or related techniques such as zero-shot, few-shot, or low-shot learning approaches.
    \item \textbf{Intervention and Comparison:} Applications of Vision Language Models (VLMs) and comparable techniques in urban environments, with comparisons between different models and approaches when available.
    \item \textbf{Outcomes:} The primary outcomes of interest include the effectiveness, accuracy, performance, and capabilities of VLMs and related techniques in urban settings. Secondary outcomes focus on advancements and future perspectives in applying these models to urban contexts.
\end{itemize}

\subsection{Stage 1: Identifying the Eligibility Criteria}\label{subsec21}

The following eligibility criteria were defined:

\begin{itemize}
    \item Documents published in English;
    \item Documents implementing either Zero-shot, Few-shot, or Vision Language Model (VLM) approaches;
    \item Data or datasets used must be from images or video, including street view and standard sensors and image formats (remote sensing excluded);
    \item Focus on objects or situations related to urban environments (urban planning, traffic, disturbances, etc.);
    \item Studies must report evaluation metrics results.
\end{itemize}

\subsection{Stage 2: Identifying the Information Sources}\label{subsec22}

The study was carried out to cover the period from January 2021 to October 2025. Due to the defined timeline, we selected specific databases as shown in Table~\ref{tab:info_sources} that would yield relevant results.

\begin{table}[h!]
\caption{Information sources used for the search phase.}\label{tab:info_sources}
\begin{tabular}{@{}lll@{}}
\toprule
\textbf{Data Source} & \textbf{Type} & \textbf{URL} \\
\midrule
Scopus & Digital library & https://www.scopus.com/ \\
DOAJ & Digital library & https://doaj.org/ \\
IEEE Xplore & Digital library & https://ieeexplore.ieee.org/ \\
ACM Digital Library & Digital library & https://dl.acm.org/ \\
JSTOR & Digital library & https://www.jstor.org/ \\
Science Direct & Digital library & https://www.sciencedirect.com/ \\
\bottomrule
\end{tabular}
\end{table}

Additionally, some targeted searches on Google Scholar were conducted to obtain specific supplementary information. These information sources were selected due to their comprehensive coverage of computer science and artificial intelligence research, as well as their ability to use complex search algorithms with logical operators that are useful in extracting the desired information.

\subsection{Stage 3: Search Strategy}\label{subsec23}

Some documents were initially read to gain a perception of the topic and formulate effective queries for the chosen libraries. Several query combinations were evaluated; however, we selected one primary search string as it yielded the most relevant results for our focus on VLMs applied to urban contexts.

With the query \texttt{("zero shot" OR "few shot" OR "low shot" OR "VLM*" OR "Vision Language Model*") AND ("urban" OR "street view")}, specific filters were applied to each information source to limit and refine the results. Since the query was identical across all databases, the main differences were in the filters applied, as detailed in Table \ref{tab:search_query}.

\begin{table}[h!]
\caption{Search queries and filters applied to each database.}\label{tab:search_query}
\begin{tabular}{@{}ll@{}}
\toprule
\textbf{Data Source} & \textbf{Filters} \\
\midrule
Scopus & Title/Keywords/Abstract, PubYear=2021-2025, Article/Conference Paper \\
DOAJ & Title/Keywords/Abstract, PubYear=2021-2025, Article/Conference Paper \\
IEEE Xplore & Title/Keywords/Abstract, PubYear=2021-2025, Article/Conference Paper \\
ACM Digital Library & Abstract only, PubYear=2021-2025, Article/Conference Paper \\
JSTOR & Title/Keywords/Abstract, PubYear=2021-2025, Article/Conference Paper \\
Science Direct & Title/Keywords/Abstract, PubYear=2021-2025, Article/Conference Paper \\
\bottomrule
\end{tabular}
\end{table}

The search was limited to recent literature (2021-2025) to capture the most current developments in VLM technology applied to urban contexts. Document types were restricted to peer-reviewed articles and conference papers to ensure scientific rigor. For most databases, searches were conducted across title, abstract, and keywords, with the exception of ACM Digital Library where only abstract searches were available with our institutional access.

\subsection{Stage 4: Selection Process}\label{subsec24}

The search process was conducted using the query algorithm in Table \ref{tab:search_query} in the digital libraries. The documents were initially screened based on title, abstract, and keywords.

Documents were examined to identify and remove duplicates, after which the remaining unique documents underwent preliminary assessment based on abstract. Documents passing this initial screening were then approved for full-text review, where a more detailed evaluation was performed to assess content relevance and quality. Each document was independently assessed by both researchers at every filtering stage, with decisions about inclusion or exclusion documented systematically. In cases of disagreement, the researchers conducted a joint re-screening of the document in question, followed by a structured debate to reach consensus on its validity and relevance to the study.

\subsection{Stage 5: Data Collection Process}\label{subsec25}

The initial search across all databases returned a total of 275 documents. After removal of duplicates, the collection was reduced to 213 unique documents.

Each document's abstract was reviewed against the eligibility criteria, resulting in 72 documents approved for full-text review. Following detailed reading of these approved documents to assess content relevance and quality, the final selection was narrowed to 32 documents that form the basis of this systematic review.

The search protocol used in this systematic review offers certain advantages. Primarily, it allowed for efficient identification of relevant research on VLMs and related techniques in urban contexts, enabling a comprehensive overview of current technologies and state-of-the-art approaches. However, limitations of this methodology include potential keyword bias due to evolving terminology in this rapidly developing field. Furthermore, while we focused on peer-reviewed publications, valuable insights from preprints or industrial reports might have been excluded.

\section{Results}\label{sec3}

\subsection{Systematic Review Protocol}\label{subsec31}

\begin{figure}[h!]
\centering
\includegraphics[width=0.8\textwidth]{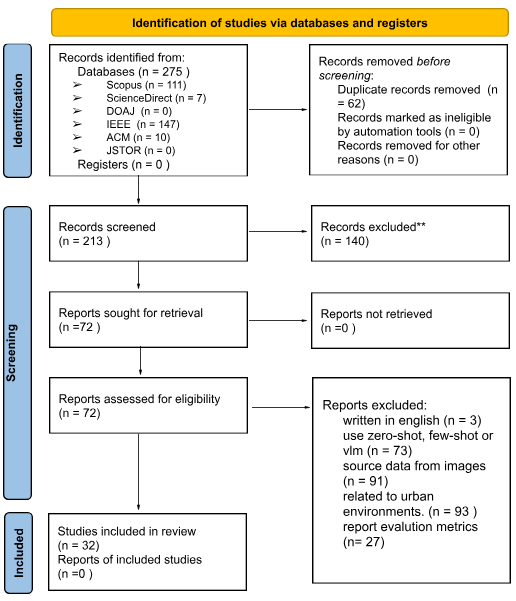}
\caption{Flowchart of the review protocol}\label{fig:flowchart}
\end{figure}

Initially, 275 documents were collected from the chosen digital libraries. These documents were distributed as follows: 111 from Scopus; 147 from IEEE Xplore; 10 from Science Direct. In the first place, 62 duplicate documents were excluded. The abstract was then read to verify whether a document met the eligibility criteria excluding 140 documents out of a total of 213, resulting in 72 documents approved for full text review. Following detailed reading of these approved documents to assess the relevance and quality of the content, the final selection was narrowed to 32 documents that form the basis of this systematic review. Fig. \ref{fig:flowchart} summarizes all the procedures of this systematic review.

\subsection{A Functional Taxonomy of VLM Applications in Urban Contexts}\label{subsec32}

\begin{figure}[h!]
    \centerline{
    \includegraphics[width=1.3\textwidth]{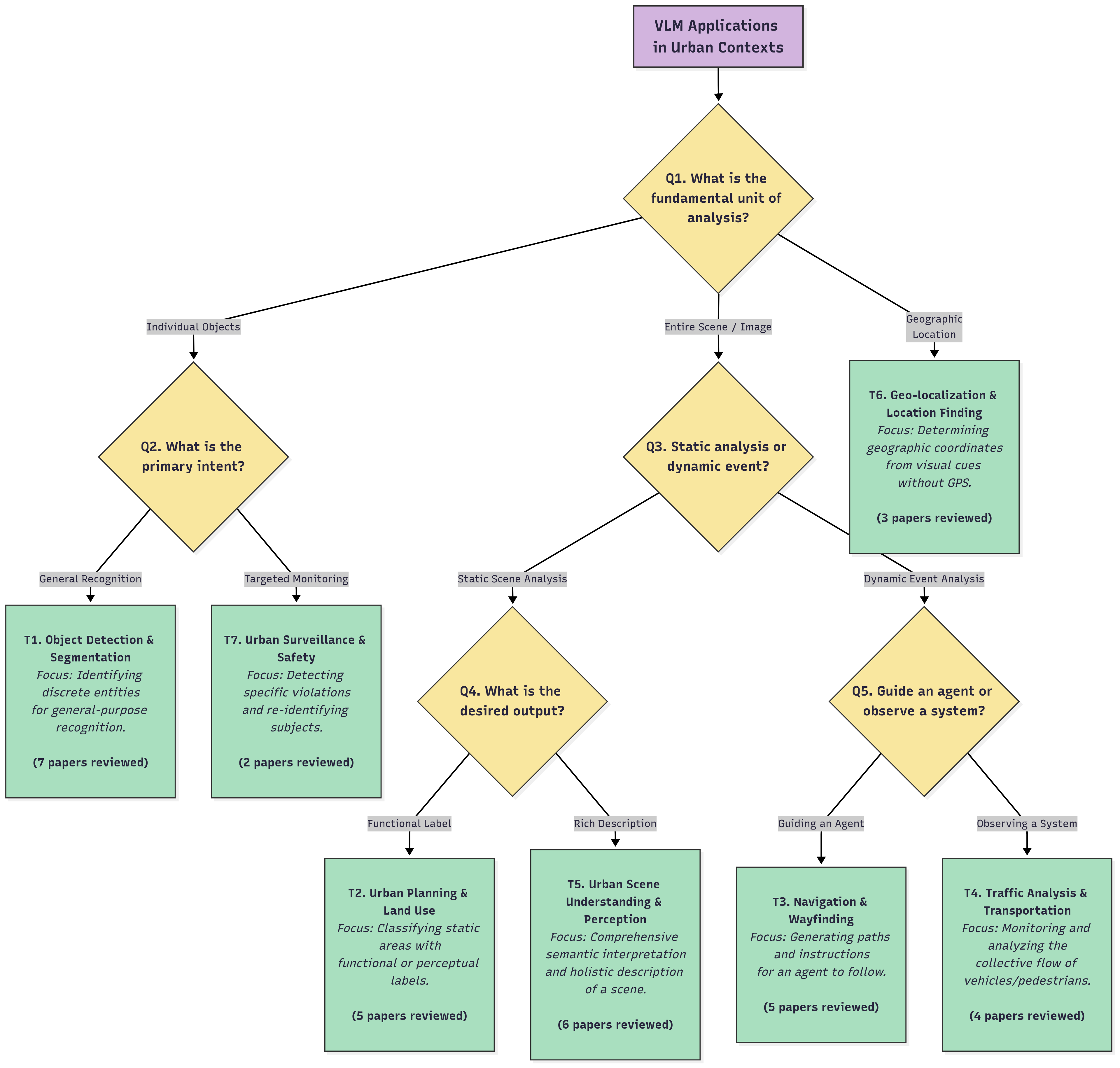} }
    \caption{A functional taxonomy of VLM applications in urban contexts. The framework categorizes the 32 reviewed studies into seven key domains based on their primary research goal. The references for studies associated with a node are listed beneath the respective node.}
    \label{fig:taxonomy_tree}
\end{figure}

The 32 papers selected for this review cover a diverse range of applications. To organize this landscape and provide a clear framework for researchers, we introduce a functional taxonomy that categorizes existing research into seven distinct domains based on its primary objective. This taxonomy is illustrated in Figure \ref{fig:taxonomy_tree}.

This framework is designed to serve as a navigational tool, enabling researchers to quickly grasp the landscape of existing approaches and identify the most suitable methods for their specific tasks. We first provide a decision guide for navigating the taxonomy, followed by a detailed discussion of each of the seven categories, summarizing the key papers and their findings.

\subsubsection{Navigating the Taxonomy}

To help researchers navigate our taxonomy, we provide a decision guide that poses specific questions at each potential juncture. By engaging with these questions, users are encouraged to reflect on their primary objectives, analytical scale, and desired output, ultimately leading them to the category that best aligns with their intended application.

\begin{itemize}

\item \textbf{Q1. What is the fundamental unit of analysis for your task?}
This is the most crucial decision. Are you analyzing discrete, countable entities within an image, or are you interpreting the image as a whole?

\begin{itemize}
    \item \textbf{Individual Objects:} If your goal is to locate, classify, or track specific items (e.g., cars, pedestrians, signs), proceed to \textbf{Q2}.
    \item \textbf{Entire Scene/Image:} If your goal is to understand the context of the entire image or video frame, proceed to \textbf{Q3}.
    \item \textbf{Geographic Location:} If your goal is to derive geographic coordinates from the visual data, your work falls into \textbf{T6. Geo-localization \& Location Finding}. This is a highly specialized task with a distinct output.
\end{itemize}

\item \textbf{Q2. (If analyzing objects) What is the primary \textit{intent} of the analysis?} The purpose behind object identification is a key differentiator. Is it for general-purpose inventory and labeling, or is it for monitoring specific behaviors or states?

\begin{itemize}
    \item \textbf{General Recognition:} If the goal is to create a general-purpose system for detecting or segmenting any object based on a prompt (e.g., "find all fire hydrants"), this aligns with \textbf{T1. Object Detection \& Segmentation}. The focus is on recognition capability.
    \item \textbf{Targeted Monitoring:} If the goal is to detect specific, predefined events or track identities for safety or security purposes (e.g., detecting helmet violations, re-identifying a person across cameras), this falls under \textbf{T7. Urban Surveillance \& Safety}. The focus is on anomaly detection and tracking.
\end{itemize}

\item \textbf{Q3. (If analyzing scenes) Is the primary goal to classify a static scene or analyze a dynamic event?} This question separates the analysis of fixed urban characteristics from the interpretation of events that unfold over time.

\begin{itemize}
    \item \textbf{Static Scene Analysis:} If one is assigning a label to a scene or generating a comprehensive description based on its persistent qualities (e.g., land use, architecture), proceed to \textbf{Q4}.
    \item \textbf{Dynamic Event Analysis:} If you are analyzing movement, interaction, or changes over time (e.g., traffic flow, an agent's journey), proceed to \textbf{Q5}.
\end{itemize}

\item \textbf{Q4. (If analyzing static scenes) What is the nature of the desired output?} The type of label or description you seek further refines the category.

\begin{itemize}
    \item \textbf{Functional Label:} If the output is a single categorical label or a score representing the area's function or quality (e.g., "commercial," "residential," "walkability score: 8/10"), your work fits into \textbf{T2. Urban Planning \& Land Use Classification}.
    \item \textbf{Rich Semantic Description:} If the output is a comprehensive, detailed description of the scene, its components, and their relationships (e.g., generating a full caption, performing dense segmentation of all elements), this aligns with \textbf{T5. Urban Scene Understanding \& Perception}.
\end{itemize}

\item \textbf{Q5. (If analyzing dynamic events) Is the system's objective to \textit{guide an agent} or to \textit{observe a system}?}
This final question distinguishes between participatory and observational dynamic tasks.

\begin{itemize}
    \item \textbf{Guiding an Agent:} If the goal is to generate a route or a set of actions for a single agent (human or robot) to follow, this is the core of \textbf{T3. Navigation \& Wayfinding}.
    \item \textbf{Observing a System:} If the goal is to analyze the collective behavior of multiple agents from an external perspective (e.g., measuring traffic density, describing a traffic scenario), this belongs to \textbf{T4. Traffic Analysis \& Transportation}.
\end{itemize}

\end{itemize}

\subsubsection{T1. Object Detection \& Segmentation}\label{subsubsec331}

Object detection and segmentation form the largest application category in our review with seven papers. These studies, categorized under task \textbf{T1}, focus on identifying specific objects in urban scenes, serving as foundation for downstream applications.

The papers employ diverse technical approaches. \cite{p2} combines Grounding DINO for bounding boxes with SAM for segmentation masks. \cite{p3} introduces VLPD with Vision-Language Semantic segmentation for contextual masks and Prototypical Semantic Contrastive learning for improved pedestrian discrimination. \cite{p5} proposes DHS-FSOD, a few-shot detection model using Dual Attention and Dynamic Hard Soft triplet loss for waste detection. \cite{p6} develops a three-module approach with YOLO-World for pseudo-annotations. \cite{p4} leverages CARLA simulator for synthetic data generation to enable real-world generalization. \cite{psingle} employs cyclic-disentangled self-distillation for domain-invariant representations across weather conditions. \cite{pfew} implements few-shot Faster R-CNN for classifying urban problems from citizen reports.

Datasets vary by application focus. CityPersons and Caltech Pedestrian are used for pedestrian detection in \cite{p3}, Huawei Waste Dataset for waste recognition \cite{p5}, FishEye8K for distorted images \cite{p6}, CARLA-generated synthetic data \cite{p4}, multi-domain weather datasets from BDD100K and FoggyCityscapes \cite{psingle}, and JAKI app images for detection of reported urban problems \cite{pfew}.

Performance results demonstrate high variability due to the different subtasks, datasets and methods being studied. The method explored in \cite{p2} achieved higher IoU scores for structured objects (cars: 0.78) than amorphous ones (trees: 0.50). \cite{p3} reports state-of-the-art performance in heavy occlusion scenarios for pedestrian detection. \cite{p4}'s CARLA-trained model outperformed 20-shot (few-shot) real-world models (11.7\% vs 1.2\% AP). The custom DHS-FSOD method from \cite{p5} achieved 32.82\% mAP50, while \cite{p6} reached an F1 score of 0.6194 in the AIC24 challenge. \cite{psingle} demonstrated strong domain generalization (56.1\% mAP on source domain, 36.6\% on night-sunny). \cite{pfew} achieved 42.5\% AP at 30-shot overall, with 69.8\% AP on novel urban problem classes.

Table \ref{tab:object-detection} summarizes these key papers in object detection and segmentation for urban contexts.

\begin{table}[h!]
\caption{Papers on Object Detection \& Segmentation}\label{tab:object-detection}
\begin{tabular}{p{0.3\textwidth}p{0.2\textwidth}p{0.2\textwidth}p{0.25\textwidth}}
\toprule
\textbf{Article} & \textbf{Method} & \textbf{Data} & \textbf{Results} \\
\midrule
Zero-Shot Object Detection and Segmentation: A Focus on Street View Imagery \cite{p2} & Combined SAM and Grounding DINO (Grounded SAM) & Street view images & IoU scores: cars (0.78), lampposts (0.78), traffic signs (0.66), trees (0.50) \\
\midrule
VLPD: Context-Aware Pedestrian Detection via Vision-Language Semantic Self-Supervision \cite{p3} & VLS segmentation and PSC learning & Caltech Pedestrian, CityPersons & 9.4\% $MR^{-2}$ (Reasonable) on CityPersons; 2.3\% $MR^{-2}$ (Reasonable) on Caltech \\
\midrule
Few-Shot Waste Detection Based on Dual Attention and Dynamic Hard Sample Triplet Loss \cite{p5} & DHS-FSOD w/ deformable convolutions (ResNet50(DCN V2)) & Huawei Waste Dataset (29 categories), MS COCO & 20.07\% mAP, 32.82\% mAP50, 20.81\% mAP75 on the Huawei Waste Dataset \\
\midrule
Improving Object Detection to Fisheye Cameras with Open-Vocabulary Pseudo-Label Approach \cite{p6} & Three-module approach w/ YOLO-World & FishEye8K (8,000 images) & F1 score of 0.6194, ranked 3rd in AIC24 Track 4 \\
\midrule
CARLA Simulated Data for Rare Road Object Detection \cite{p4} & Synthetic data generation w/ CARLA \& ResNet 101 for detection & CARLA synthetic, MS COCO, MVD & (CARLA) Detector trained with domain-randomized synthetic data on the MVD test set: (11.7\% AP) crosswalk detection; (37.5\% AP) fire hydrant detection \\
\midrule
Single-Domain Generalized Object Detection in Urban Scene via Cyclic-Disentangled Self-Distillation \cite{psingle} & Cyclic-disentangled self-distillation with Faster R-CNN & 32,153 daytime, night and rainy scenes from BDD100K; 3,755 foggy scenes from FoggyCityscapes and Adverse-Weather & 56.1\% mAP on source domain; 36.6\% mAP on night-sunny; 28.2\% mAP on dusk-rainy \\
\midrule
Few-Shot Object Detection for Classifying Reported Urban Problems \cite{pfew} & Faster R-CNN with ResNet50 FPN 3x backbone in few-shot setting & 2,402 images of urban problems from Jakarta citizen reporting app (JAKI) & 42.5\% AP at 30-shot for all classes; 69.8\% AP for novel classes \\
\bottomrule
\end{tabular}
\end{table}

\subsubsection{T2. Urban Planning \& Land Use Classification}\label{subsubsec332}

Task \textbf{T2} of our taxonomy includes five papers that address urban planning and land usage classification, using vision-language integration to interpret complex urban functions. These papers focus on higher-level semantic concepts like mixed land use \cite{p7}, walkability \cite{p8}, and urban functions \cite{p9}.

Technical approaches are varied with prompt engineering being a common strategy. \cite{p7} uses CLIP with spatial context-aware prompts incorporating location information to improve land use classification. \cite{p8} creates a system for assessing walkability using custom CLIP prompts with entropy adjustment to balance positive and negative urban indicators. \cite{p9} introduces UrbanCLIP, mapping abstract urban functions to 354 concrete Urban Object Types through specialized templates. \cite{p10} implements a deep semantic-aware network (DSANZS) with a semantic correlation matrix for zero-shot urban perception. \cite{p11} combines visual features from Grounding DINO and DINOv2 with text extraction using CLIPSeg, CRAFT, PARSeq and GPT-3.5.

All papers primarily use street-view imagery: Google Street View \cite{p7}\cite{p11}, Mapillary \cite{p8}, and Baidu Map SVIs \cite{p9}. Sample sizes range from 3,398 images \cite{p7} to 226,881 images \cite{p9}. For urban perception, \cite{p10} used a Place Pulse 2.0 subset with 104,529 images, while \cite{p11} collected 539 SVIs from four Indian cities for zoning or usage prediction.

Performance metrics show strong zero-shot results. \cite{p7} achieved 66.39\% matching accuracy with context-aware prompts. \cite{p9} reached F1 scores of 0.82 for residential areas and 0.72 for commercial zones. Mixed-use areas showed higher accuracy (71.18\%) than single-function areas (62.67\%) in \cite{p7}. For urban perception, \cite{p10} demonstrated improved accuracy across all attributes, reaching 69.20\% for beauty perception. The ensemble model in \cite{p11} achieved a macro average F1 of 0.84, significantly improving commercial and institutional building classification compared to vision-only approaches.

Table \ref{tab:urban-planning} summarizes the key papers in this category.

\begin{table}[h!]
\caption{Papers on Urban Planning and Land Use Classification}\label{tab:urban-planning}
\begin{tabular}{p{0.35\textwidth}p{0.2\textwidth}p{0.2\textwidth}p{0.2\textwidth}}
\toprule
\textbf{Article} & \textbf{Method} & \textbf{Data} & \textbf{Results} \\
\midrule
Mixed land use measurement and mapping with street view images and spatial context-aware prompts via zero-shot multimodal learning \cite{p7} & CLIP with spatial context-aware prompts and ensemble approach & 3,398 GSV images from NYC (Place Plus 2.0 subset), OpenStreetMap data w/ 6 urban planning categories & 64.56\% matching with prompt ensembling (50m); 66.39\% w/ context-aware prompts (50m) \\
\midrule
A New Approach to Assessing Perceived Walkability: Combining Street View Images with Multimodal Contrastive Learning Model \cite{p8} & CLIP with custom walkability prompt scale and entropy adjustment & 5,669 Mapillary images from Amsterdam Centrum w/ a 30m interval & Identified walkability hotspots missed by objective metrics \\
\midrule
Zero-shot urban function inference with street view images through prompting a pretrained vision-language model \cite{p9} & UrbanCLIP framework with urban taxonomy and 6 specialized templates &  226,881 SVIs from Shenzhen, China (Baidu Map API); annotated subset of 1,518 images (1,179 single-label, 339 multi-label) & 0.655 weighted F1 score on the primary function classification subset; 0.696 weighted F1 score on the multiple function classification \\
\midrule
Deep semantic-aware network for zero-shot visual urban perception \cite{p10} & Two-stream network with Semantic Correlation Matrix for zero-shot learning (DSANZS) & Place Pulse 2.0 subset (104,529 images) & Improved accuracy across all categories against all other baselines on the PP 2.0 subset \\
\midrule
Building usage prediction in complex urban scenes by fusing text and facade features from street view images using deep learning \cite{p11} & Vision-language ensemble with Grounding DINO, DINOv2 for visual feature extraction; CLIPSeg, CRAFT, PARSeq and GPT-3.5 for text feature extraction & 539 SVIs from four Indian cities & 0.84 ensemble macro average F1 \\
\bottomrule
\end{tabular}
\end{table}

\subsubsection{T3. Navigation \& Wayfinding}\label{subsubsec333}

The third category, \textbf{T3}, comprises five papers that address navigation challenges in urban environments using vision-language integration, focusing on natural language instruction following, POI (Point of Interest) identification, and path planning.

The papers present varied technical approaches. \cite{p12} introduces a three-stage pipeline for POI data generation combining Cascade Mask R-CNN for region identification with DB/SRN for text detection and a Visual-Linguistic Multi-task Classification model for ROI classification. \cite{p13} presents a modular vehicle navigation system with ChatGLM for landmark extraction from instructions, using CLIP and BLIP-2 for matching environmental images with landmarks. \cite{p14} introduces PM-VLN with a Priority Map module that conducts hierarchical trajectory estimation through ConvNeXt Tiny backbone and feature-level localization. \cite{p15} develops a hierarchical few-shot waypoint detection system using a ResNet-50 backbone enhanced with distribution embedding and symmetrized Mahalanobis distance metric. \cite{p16} implements ViPlanner, an end-to-end learned multi-domain planner combining geometric and semantic information through perception/planning networks and Bi-Level Optimization.

Datasets vary widely across papers. \cite{p12} employed approximately 2.7 million street view images from Shenzhen, including Baidu Street View (1.46M images) and Tencent Street View (1.24M images). \cite{p13} relied on Google Maps Street View with 360° camera views. \cite{p14} uses the Touchdown benchmark (6,525 Manhattan routes with natural language instructions). \cite{p15} created a custom dataset covering 36 courses/paths across 11 UC Davis buildings. \cite{p16} used NVIDIA Omniverse with environments from Matterport3D, CARLA, and custom warehouse settings to generate approximately 80,000 start-goal pairs.

Performance metrics demonstrate strong results across navigation tasks. \cite{p12} achieved an F1-score of 84.23\% for ROI classification, generating 815,616 POI records. \cite{p13} reached 46.5\% success rate with optimal camera setup, outperforming traditional object detection models by 14.6\%. \cite{p14}'s Priority Map approach achieved 33.4\% task completion on the Touchdown benchmark. \cite{p15} successfully demonstrated robust waypoint detection using minimal training examples, with their symmetrized Mahalanobis metric outperforming other distance metrics. \cite{p16} demonstrated a 38.02\% semantic loss decrease compared to geometric-only approaches, with particularly strong performance in urban navigation where the system correctly commanded robots to use crosswalks and prefer sidewalks over roads.

Table \ref{tab:navigation} summarizes the key papers in this category.

\begin{table}[h!]
\caption{Papers on Navigation \& Wayfinding}\label{tab:navigation}
\begin{tabular}{p{0.35\textwidth}p{0.2\textwidth}p{0.2\textwidth}p{0.2\textwidth}}
\toprule
\textbf{Article} & \textbf{Method} & \textbf{Data} & \textbf{Results} \\
\midrule
Deep-learning generation of POI data with scene images \cite{p12} & Three-stage pipeline with Cascade Mask R-CNN, DB/SRN, and VLMC & BSV (1.46M images) and TSV (1.24M images); Four STR datasets (ArT, RCTW-17, LSVT, ReCTS); Custom datasets: 5,436 images for ROI segmentation, 5,553 samples for classification & F1-score of 84.23\% for ROI classification; Generated 815,616 POI records with 70.94\% overlap with existing databases \\
\midrule
Enabling Vision-and-Language Navigation for Intelligent Connected Vehicles Using Large Pre-Trained Models \cite{p13} & Three-component system with ChatGLM, CLIP, and BLIP-2 & Google Street View data, 360° images from 8 cameras & 14.6\% higher success rate than baseline; SR=46.5\% with optimal camera setup \\
\midrule
A Priority Map for Vision-and-Language Navigation with Trajectory Plans and Feature-Location Cues \cite{p14} & PM-VLN with trajectory estimation and feature localization & Touchdown (6,525 routes), TR-NY-PIT-central (17,000 routes), MC-10 (8,100 landmarks) & 33.4\% Task Completion on test set \\
\midrule
Hierarchical End-to-End Autonomous Navigation Through Few-Shot Waypoint Detection \cite{p15} & Few-shot waypoint detection with distribution embedding & 36 courses/paths across 11 UC Davis buildings recorded with fisheye cameras & Successful navigation with minimal examples \\
\midrule
ViPlanner: Visual Semantic Imperative Learning for Local Navigation \cite{p16} & End-to-end planner with semantic understanding and Bi-Level Optimization & NVIDIA Omniverse with Matterport3D, CARLA, and warehouse environments & 38.02\% semantic loss decrease in CARLA and warehouse environment; successful zero-shot transfer to real-world \\
\bottomrule
\end{tabular}
\end{table}

\subsubsection{T4. Traffic Analysis \& Transportation}\label{subsubsec334}

Task \textbf{T4} contains four papers that focus on analyzing traffic patterns, vehicle tracking, and transportation infrastructure, enhancing monitoring systems for safety assessment and autonomous driving perception.

The papers explore different approaches targeting different aspects of traffic analysis. \cite{p17} introduces a two-stage traffic description approach combining three captioning architectures (single-view, motion-blur, and multi-view) with a rule-based refinement engine featuring pedestrian-aware, vehicle-aware, and context-aware modules. \cite{p18} presents a segment-based methodology breaking down complex traffic descriptions into five semantic components (appearance, environment, location, attention, and action) using a two-phase extraction pipeline with Mistral 7B and MiniLM. \cite{p19} enhances Grounded-SAM for autonomous driving with parameter optimization (box threshold 0.30, text threshold 0.25), CLIP-based verification, and a specialized ResNet34 model for traffic sign recognition. \cite{p20} implements an MLVR system integrating five modules: X-CLIP-based video recognition, Tip-Adapter-based vehicle attribute classification, trajectory-based motion analysis, surrounding vehicle detection, and a match control system for optimization.

The datasets used in this task are more standardized, with some paper even focusing on a specific challenge. \cite{p17} and \cite{p18} are focused on Track 2 of the AIC24 and both use the WTS (Woven Traffic Safety) Dataset containing over 1,200 internal videos spanning 130 unique traffic scenarios and 4,800+ external videos from the BDD100K dataset. \cite{p19} utilized BDD100K with approximately 100,000 videos from four US cities (New York, Berkeley, San Francisco, Bay Area) and the GTSRB dataset with 43 traffic sign classes for specialized sign classification. \cite{p20} used CityFlow-NL with 2,155 vehicle trajectories and corresponding natural language descriptions, plus a test set of 184 distinct vehicle trajectories.

Performance results demonstrate the effectiveness of these approaches for traffic analysis. \cite{p17}'s best configuration (motion-blur model with BLIP-2) ranked 9th with a score of 22.74 in the AIC24 Track 2, with significant improvement from rule-based refinement. \cite{p18} ranked 2nd with a score of 32.89, demonstrating a 60\% improvement with dynamic prompting and segment-specific adapters. \cite{p19} achieved 95\% accuracy on traffic sign classification with its ResNet34 model, showing qualitative improvements in vehicle type differentiation and sign recognition compared to the baseline Grounded-SAM. \cite{p20} achieved 81.79\% Mean Reciprocal Rank accuracy through its multi-module approach, with ablation studies showing the vehicle motion module (+16.94\%) and surrounding module (+10.19\%) providing the largest performance gains.

Table \ref{tab:traffic} summarizes the key papers in this category.

\begin{table}[h!]
\caption{Papers on Traffic Analysis \& Transportation}\label{tab:traffic}
\begin{tabular}{p{0.35\textwidth}p{0.2\textwidth}p{0.2\textwidth}p{0.2\textwidth}}
\toprule
\textbf{Article} & \textbf{Method} & \textbf{Data} & \textbf{Results} \\
\midrule
Multi-perspective Traffic Video Description Model with Fine-grained Refinement Approach \cite{p17} & Two-stage approach with multi-perspective model with three submodels (single-view, motion-blue and multi-view) and rule-based refinement & Woven Traffic Safety (WTS) dataset (1,200+ video events); BDD100K subset (detailed textual annotations for 4,800+ videos)  & Ranked 9th in AIC24 Track 2 with score of 22.74; best performance with motion-blur and BLIP-2 \\
\midrule
Divide and Conquer Boosting for Enhanced Traffic Safety Description and Analysis with Large Vision Language Model \cite{p18} & Segment-based approach with two-phase extraction using Qwen-VL & WTS dataset (1,200+ video events); BDD100K subset (detailed textual annotations for 4,800+ videos) & Ranked 2nd in AIC24 Track 2 with score of 32.89; 60\% improvement with dynamic prompting \\
\midrule
Enhancing Autonomous Driving with Grounded-Segment Anything Model: Limitations and Mitigations \cite{p19} & Grounded-SAM-ZP with CLIP verification and ResNet34 for sign classification & BDD100K (100K videos), GTSRB (43 sign classes) & 95\% accuracy on BDD100K subset; Significant qualitative improvements in vehicle type differentiation and sign recognition \\
\midrule
A Unified Multimodal Structure for Retrieving Tracked Vehicles through Natural Language Descriptions \cite{p20} & Five-module MLVR system with X-CLIP adaptation and Tip-Adapter & CityFlow-NL (2,155 vehicle trajectories); Custom evaluation set with 184 distinct vehicle trajectories & 81.79\% MRR accuracy; Ranked 2nd in AIC23 Track 2 \\
\bottomrule
\end{tabular}
\end{table}

\subsubsection{T5. Urban Scene Understanding \& Perception}\label{subsubsec335}

The fifth category in our taxonomy, \textbf{T5}, includes six papers focused on the comprehensive understanding of urban scenes through semantic segmentation, scene parsing, and holistic interpretation. Unlike object detection which targets specific elements, these works develop complete scene understanding with emphasis on domain adaptation across different urban contexts.

The papers implement diverse technical approaches leveraging various VLM architectures. \cite{p21} introduces CityLLaVA, an efficient fine-tuning framework built on LLaVA-1.6-34B that employs "block expansion" instead of LoRA, adding visual prompting through colored rectangles (green for pedestrians, blue for vehicles) and global-local joint views. \cite{p22} presents CBFA, reinterpreting few-shot cross-domain scene parsing through Rubin's causal inference framework using DeepLabV2 with ResNet-101 backbone. \cite{p23} develops SemiMTL with a ResNet101 encoder and domain-aware discriminators for cross-dataset learning through adversarial training. \cite{p24} implements CLIPTER, integrating scene context into text recognition by combining pretrained CLIP features with multi-head cross-attention into existing text recognition models. \cite{p25} creates a self-supervised depth estimation approach using ConvNeXt-B backbone with aspect ratio augmentation for better generalization. \cite{p1} introduces AZVG, a zero-shot video grounding pipeline using CLIP with three modules: PSG (Prompting Sentences Generation) for query enhancement, CAG (Candidate Anchors Generation) for identifying relevant frames, and ATPD (Atom-based Time Period Detection) for temporal segment detection.

The datasets demonstrate diverse urban environments. \cite{p21} is focused on the Track 2 of the AIC24 and uses the same WTS and BDD100K dataset from \cite{p17} and \cite{p18}. The method explored in this paper, however, is transferable to tasks outside of traffic analysis justifying it's placement in this task. This paper also introduces a custom image-text dataset with long question-answer and short question-answer pairs. \cite{p22} evaluates domain adaptation between synthetic datasets (GTA5 with 24K images, SYNTHIA with 9.4K images) and real-world CITYSCAPES (2,975 images from 18 cities). \cite{p23} conducts experiments across three scenarios with varying domain gaps using Cityscapes, Potsdam, Vaihingen, and Synscapes datasets. \cite{p24} uses 12 text recognition datasets including street-view collections like SVT, LSVT, and Uber (totaling over 200K test words). \cite{p25} introduces the SlowTV dataset with 1.7M frames from 135 hours of YouTube videos covering hiking, driving, and underwater environments. \cite{p1} evaluates on Charades-STA (6,672 videos with 16,128 moment-sentence pairs) and ActivityNet Captions (19,209 videos with 72,000 moment-sentence pairs).

Performance results demonstrate the effectiveness of these approaches. \cite{p21}'s CityLLaVA achieved a benchmark score of 33.43 on the 2024 AI City Challenge Track 2 (first place), with global+local view combination proving most effective. \cite{p22}'s CBFA reached 57.52\% mIoU on GTA5 (CITYSCAPES) with just 5 target images, outperforming state-of-the-art by 9.34\%. \cite{p23}'s SemiMTL showed 22.5\% mIoU improvement for segmentation without any segmentation labels using their domain-aware discriminators. \cite{p24}'s CLIPTER reduced relative error by nearly 10\% on street-view datasets, with a 0.8\% weighted average improvement when integrated with PARSeq. \cite{p25} outperformed all self-supervised baselines in zero-shot generalization through its aspect ratio augmentation and learned camera intrinsics. \cite{p1}'s AZVG achieved 39.01\% R@1, IoU=0.5 on Charades-STA without any training, significantly outperforming the supervised CTRL method (23.63\%).

A key distinction from other categories is the emphasis on domain adaptation, with all six papers addressing the challenge of transferring knowledge across different urban environments with minimal supervision.

Table \ref{tab:scene-understanding} summarizes the key papers in this category.

\begin{table}[h!]
\caption{Papers on Urban Scene Understanding \& Perception}\label{tab:scene-understanding}
\begin{tabular}{p{0.35\textwidth}p{0.2\textwidth}p{0.2\textwidth}p{0.2\textwidth}}
\toprule
\textbf{Article} & \textbf{Method} & \textbf{Data} & \textbf{Results} \\
\midrule
CityLLaVA: Efficient Fine-Tuning for VLMs in City Scenario \cite{p21} & Block expansion fine-tuning of LLaVA-1.6-34B with colored visual prompting and sequential questioning & WTS dataset (1,200+ video events); BDD100K subset (detailed textual annotations for 4,800+ videos); Custom image-text dataset with long and short QA pairs & Ranked 1st in
AIC24 Track 2 with
score of 33.43 \\
\midrule
Counterfactual Balancing Feature Alignment for Few-Shot Cross-Domain Scene Parsing \cite{p22} & CBFA with DeepLabV2 (ResNet-101), inverse propensity score weighting and counterfactual matching & GTA5 (24K images), SYNTHIA (9.4K), CITYSCAPES & 57.52\% mIoU with 5-shot adaptation; 9.34\% improvement over joint training \\
\midrule
Semi-supervised Multi-task Learning for Semantics and Depth \cite{p23} & SemiMTL with ResNet101 backbone and domain-aware discriminators for cross-dataset learning & Cityscapes, Cityscapes-depth, Potsdam, Vaihingen, Synscapes & 22.5\% mIoU improvement on Vaihingen without segmentation labels \\
\midrule
CLIPTER: Looking at the Bigger Picture in Scene Text Recognition \cite{p24} & Context-aware text recognition with frozen CLIP for scene embeddings and multi-head cross-attention for integration & 12 datasets including SVT, LSVT, Uber (200K+ test words) & State-of-the-art results by integrating CLIPTER with PARSeq increasing its accuracy by +0.8\% across all datasets; 10\% error reduction on street-view datasets \\
\midrule
Kick Back \& Relax: Learning to Reconstruct the World by Watching SlowTV \cite{p25} & Self-supervised depth estimation with ConvNeXt-B backbone, aspect ratio augmentation, and learned camera intrinsics & SlowTV (1.7M frames), Mannequin Challenge (115K), KITTI (71K) & Outperformed all self-supervised baselines in zero-shot generalization with 61.5 FPS inference speed \\
\midrule
Zero-Shot Video Grounding for Automatic Video Understanding in Sustainable Smart Cities \cite{p1} & AZVG framework leveraging CLIP with prompting sentences generation, candidate anchors, and atom-based time period detection & Charades-STA (16,128 pairs), ActivityNet Captions (72,000 pairs) & 39.01\% R@1, IoU=0.5 on Charades-STA; 47.37\% R@1, IoU=0.3 on ActivityNet \\
\bottomrule
\end{tabular}
\end{table}

\subsubsection{T6. Geo-localization \& Location Finding}\label{subsubsec336}

Task \textbf{T6} consists of three papers focused on determining geographic location from visual inputs without relying on GPS, leveraging vision-language models to interpret distinctive visual elements that reveal geographic context.

The papers employ diverse technical approaches. \cite{p26} introduces GeoReasoner, a QwenVL-based architecture with three key components: Vision Encoder (ViT), Vision-Language Adapter, and Pre-trained LLM. The model employs a novel "locatability" metric developed with MaskFormer segmentation masks and Sentence-BERT similarity measurements to filter their training data, followed by two-stage LoRA fine-tuning: reasoning tuning with 3K textual clues from geo-localization games, then location tuning using 70K highly locatable GSV images. \cite{p27} presents two GEM models: Zero-shot GEM applying geo-contextualized prompts to CLIP (e.g., "in \{city\}"), and Linear-probing GEM combining CLIP's ViT-L/14 encoder with a logistic regression classifier. \cite{p28} implements AdAGeo, a two-phase architecture: first, a few-shot domain-driven data augmentation (DDDA) module using dual parallel autoencoders with reconstruction, cycle-consistency, and variational losses to learn source-to-target domain mapping; second, a visual place recognition block combining CNN features with Places365-based attention maps, a domain adaptation module with gradient reversal layer (GRL), and a NetVLAD aggregator trained with weakly supervised triplet margin loss.

Dataset selection relies heavily on street-view imagery. \cite{p26} employs 70K highly locatable Google Street View images filtered from 130K initial images across 48 countries using their proposed locatability metric. \cite{p27} uses Place Plus 2.0 with 111K GSV images from 56 cities. \cite{p28} introduces SVOX, combining 74,646 GSV gallery images with Oxford RobotCar queries across five environmental conditions (Snow, Rain, Sun, Night, Overcast), with approximately 1,500 images per target domain.

Performance results demonstrate strong capabilities. \cite{p26}'s GeoReasoner achieved 82.4\% country-level and 75.2\% city-level accuracy, significantly outperforming other LVLMs like LLaVA (40.29\%) and Qwen-VL (58.29\%). \cite{p27}'s Zero-shot GEM reached 64.4\% top-1 accuracy, while their Linear-probing variant achieved 85.9\% top-1 accuracy. \cite{p28}'s AdAGeo demonstrated 49.8\% average Recall@1 with ResNet18 using only 5 target domain images, with best performance on "Overcast" (80.1\%) and most challenging on "Night" (10.5\%), outperforming other domain adaptation methods while requiring significantly fewer target images.

A distinctive feature is these model's emphasis on interpretability, with \cite{p26} providing detailed reasoning for predictions by leveraging textual clues from geo-localization games, and \cite{p27} demonstrating how language prompting focuses attention on location-relevant features.

Table \ref{tab:geo-localization} summarizes the key papers in this category.

\begin{table}[h!]
\caption{Papers on Geo-localization \& Location Finding}\label{tab:geo-localization}
\begin{tabular}{p{0.35\textwidth}p{0.2\textwidth}p{0.2\textwidth}p{0.2\textwidth}}
\toprule
\textbf{Article} & \textbf{Method} & \textbf{Data} & \textbf{Results} \\
\midrule
GeoReasoner: Geo-localization with Reasoning in Street Views using a Large Vision-Language Model \cite{p26} & QwenVL-based architecture with ViT encoder, VL-Adapter, two-stage LoRA fine-tuning, and "locatability" filtering & 70K highly locatable GSV images from 130K initial samples across 48 countries, 3K textual clues from geo-localization games & 82.4\% country-level accuracy, 75.2\% city-level accuracy; F1 scores of 0.9033 (country) and 0.8584 (city) \\
\midrule
IM2City: Image Geo-localization via Multimodal Learning \cite{p27} & Zero-shot GEM with CLIP and geo-contextualized prompting; Linear-probing GEM with CLIP ViT-L/14 encoder and logistic regression & Place Plus 2.0 (111K images from 56 cities), IM2GPS3k benchmark & Zero-shot: 64.4\% top-1 accuracy with prompt ensembling; Linear-probing: 85.9\% top-1 accuracy on Place Plus 2.0 \\
\midrule
Adaptive-Attentive Geolocalization from few queries: a hybrid approach \cite{p28} & AdAGeo: two-phase geo-localization architecture with dual autoencoders for domain adaptation with reconstruction and cycle-consistency losses & SVOX dataset: 74,646 GSV images as gallery; $\approx1,500$ images per target domain (Snow, Rain, Sun, Night, Overcast) from Oxford RobotCar & 49.8\% average Recall@1 with ResNet18 using only 5 target domain images; 80.1\% Recall@1 on Overcast domain \\
\bottomrule
\end{tabular}
\end{table}

\subsubsection{T7. Urban Surveillance \& Safety}\label{subsubsec337}

The final category, \textbf{T7}, includes two papers addressing critical aspects of urban safety through person re-identification and violation detection. Both emphasize few-shot learning approaches valuable in surveillance contexts where labeled examples of specific violations may be limited.

\cite{p29} introduces ReWIF (Re-Weighting Instance method based on influence Function), a two-stage approach that first trains a ResNet50, then fine-tunes it by weighting source samples based on their influence on a small target domain support set. \cite{p30} implements a few-shot data sampling framework with three steps: background determination through median computation, video categorization by lighting conditions (day, night, fog), and adaptive frame sampling to select more frames from underrepresented categories, combined with YOLOv8 and Test Time Augmentation.

Dataset selection reflects specialized surveillance applications. \cite{p29} employs multiple person re-ID benchmarks: Market-1501 (32,668 images of 1,501 identities from 6 cameras), DukeMTMC-reID (36,411 images of 702 identities from 8 cameras), and three CUHK datasets to simulate cross-city adaptation scenarios. \cite{p30} used data from the 2023 NVIDIA AI CITY CHALLENGE (Track 5) with 200 videos, selecting 4,500 representative frames split 70\%/30\% between training and validation.

Performance results are able to showcase good few-shot capabilities. \cite{p29}'s ReWIF achieved 82.2\% mAP and 95.9\% Rank-1 accuracy with 50 target identities, dramatically outperforming baselines (21.2\% mAP for source-only training). Even with just one identity, it maintained 67.2\% mAP. Multi-source experiments showed further improvements (85.1\% mAP). \cite{p30} reached 92.3\% mAP@.05 on validation and 58.6\% mAP on the test dataset while maintaining real-time inference (95 fps), ranking 7th in the competition.

Table \ref{tab:surveillance} summarizes the key papers in this category.

\begin{table}[h!]
\caption{Papers on Urban Surveillance \& Safety}\label{tab:surveillance}
\begin{tabular}{p{0.35\textwidth}p{0.2\textwidth}p{0.2\textwidth}p{0.2\textwidth}}
\toprule
\textbf{Article} & \textbf{Method} & \textbf{Data} & \textbf{Results} \\
\midrule
An Easy Way to Deploy the Re-ID System on Intelligent City: Cross-Domain Few-Shot Person Reidentification \cite{p29} & ReWIF with influence function-based sample weighting and two-stage fine-tuning & Market-1501 (32K images of 1,501 identities), DukeMTMC-reID (36K images of 702 identities), CUHK datasets & 82.2\% mAP, 95.9\% Rank-1 with 50 identities; 67.2\% mAP with just 1 identity; 85.1\% mAP with multi-source training \\
\midrule
Real-time Multi-Class Helmet Violation Detection Using Few-Shot Data Sampling Technique and YOLOv8 \cite{p30} & Few-shot data sampling framework with video categorization, adaptive sampling, YOLOv8 and Test Time Augmentation & 2023 NVIDIA AI CITY CHALLENGE, Track 5 (200 videos at 10 fps, 1920×1080 resolution); 4,500 selected frames & 92.3\% mAP@.05, 64.7\% mAP.05-.95 on validation; 58.6\% mAP on test with 95 fps inference speed; ranked 7th in competition \\
\bottomrule
\end{tabular}
\end{table}

\subsection{Dataset Usage Patterns}\label{sec:results_dataset_trends}

Across the 32 reviewed studies, more than 30 datasets were used to train or evaluate models in zero-shot and few-shot urban scenarios. These datasets span four primary categories: (1) structured street-level imagery, (2) synthetic simulations, (3) aerial or top-down views, and (4) proprietary, task-specific collections.

\textbf{Street-Level Imagery.}
Datasets such as \textit{Cityscapes}~\cite{p22,p23}, \textit{BDD100K}~\cite{p17,p18,p19,p21}, \textit{Google Street View (GSV)}~\cite{p7,p11,p13,p26,p27}, \textit{Mapillary Vistas}~\cite{p8}, and \textit{Baidu SVIs}~\cite{p9,p12} were among the most frequently reused resources. These datasets provide high-resolution imagery aligned with urban perception tasks including semantic segmentation, pedestrian detection, navigation, and geo-localization. While \textit{Cityscapes} and \textit{BDD100K} offer standardized annotations, broader platforms like GSV and Mapillary introduce geographic and cultural diversity but often require additional preprocessing due to inconsistent labeling.

\textbf{Synthetic and Aerial Datasets.}
Synthetic datasets such as \textit{CARLA}~\cite{p16,p25}, \textit{SYNTHIA}~\cite{p22}, and \textit{GTA5}~\cite{p22} were widely adopted to simulate rare, hazardous, or ethically sensitive conditions. These datasets enabled scalable zero-shot training and domain adaptation but introduced domain gaps requiring real-world fine-tuning. Aerial datasets like \textit{Potsdam} and \textit{Vaihingen}~\cite{p23} supported land use classification and semantic segmentation from overhead perspectives, albeit in fewer studies.

\textbf{Proprietary and Domain-Specific Data.}
Several works relied on internal or non-public datasets tailored to specific deployment needs—such as traffic violation detection~\cite{p30}, waste monitoring~\cite{p12}, and camera distortion modeling~\cite{p21}. Examples include the \textit{Huawei Waste Dataset}~\cite{p12}, \textit{FishEye8K}~\cite{p21}, and AI City Challenge videos~\cite{p17,p18,p21,p30}. While these datasets enhance relevance to real-world use cases, their lack of accessibility hinders reproducibility and community benchmarking.

\textbf{Cross-City Generalization.}
Only a minority of studies evaluated models across distinct urban contexts. Most benchmarks used data from a single city or region, limiting insights into geographic transferability. Notable exceptions include domain adaptation studies using synthetic-to-real setups such as \textit{GTA5}, and \textit{Cityscapes}~\cite{p22}. However, few studies tested on cities with differing architecture, signage systems, or socio-cultural environments.

\textbf{Temporal Robustness and Realism.}
Although video datasets as \textit{BDD100K}~\cite{p17,p18,p19,p21} are available, temporal modeling was rarely the focus. Most evaluations used isolated frames, overlooking temporal drift, motion continuity, or sequence consistency. This restricts the evaluation of models in streaming or real-time urban scenarios like driving or surveillance.

\textbf{Deployment-Oriented Evaluation.}
Evaluation metrics typically prioritized accuracy or segmentation quality, with limited consideration of real-world deployment factors. Only a few studies measured runtime, latency, or hardware constraints~\cite{p30}. Sensor realism (e.g., fisheye cameras~\cite{p21}, environmental noise), annotation cost, or ethical compliance (e.g., data provenance, fairness, demographic bias) were either qualitatively noted or omitted entirely.

Overall, while dataset diversity is improving, evaluation practices lag behind deployment needs. Few studies benchmark cross-city or multi-frame generalization, and even fewer account for operational constraints. As the field progresses, stronger alignment between dataset use and real-world applicability remains a critical challenge. A comprehensive dataset inventory is available in Appendix~\ref{secA1}, Tables~\ref{tab:dataset-real}–\ref{tab:dataset-private}.

\subsection{Model Landscape: A Functional Taxonomy}

Urban applications of VLMs are shaped by the interplay of vision, language, and multimodal architectures, each fulfilling distinct roles in processing and interpreting complex urban scenes. Based on the 32 reviewed papers, we classify the model ecosystem into four categories: vision-only backbones, standalone language models, and multimodal VLMs. This taxonomy captures the operational patterns, integration strategies, and current limitations across tasks such as object detection, planning, navigation, and geo-perception.

\textbf{Vision Models.} Vision backbones dominate urban AI pipelines due to their modularity, maturity, and efficiency. Models such as \textit{CLIP}, \textit{Grounding DINO}, and \textit{YOLOv8} are widely deployed for object localization, semantic segmentation, and visual grounding, particularly in street-level tasks. Specialized encoders like \textit{SAM}, \textit{CLIPSeg}, and \textit{DPT} are leveraged for zero-shot segmentation, monocular depth estimation, and layout inference. Despite their versatility, most are used as frozen components, often without fine-tuning, which raises concerns around adaptation in geographically or culturally diverse urban contexts.

\textbf{Language Models.} LLMs such as \textit{GPT-3.5}, and \textit{ChatGLM} play auxiliary roles in prompt generation, semantic reasoning, or interpreting structured metadata. These models are rarely fine-tuned; instead, they operate via API access or static checkpoints, limiting contextual adaptation. While they offer powerful reasoning capabilities, their use is often opaque, few studies document prompt structures, error behavior, or linguistic grounding failures.

\textbf{Multimodal Vision-Language Models.}
VLMs such as \textit{BLIP-2}, and \textit{LLaVA} provide the most direct form of joint visual-linguistic reasoning. They are increasingly used for captioning, question answering, and grounding in planning or perception workflows. However, their integration remains brittle: models are often deployed “as-is,” with minimal tuning or auditing. Tasks demanding fine-grained spatial understanding or region-specific semantics often require custom prompt engineering or multimodal fusion pipelines.

\begin{figure}[h!]
    \centering
    \includegraphics[width=0.7\linewidth]{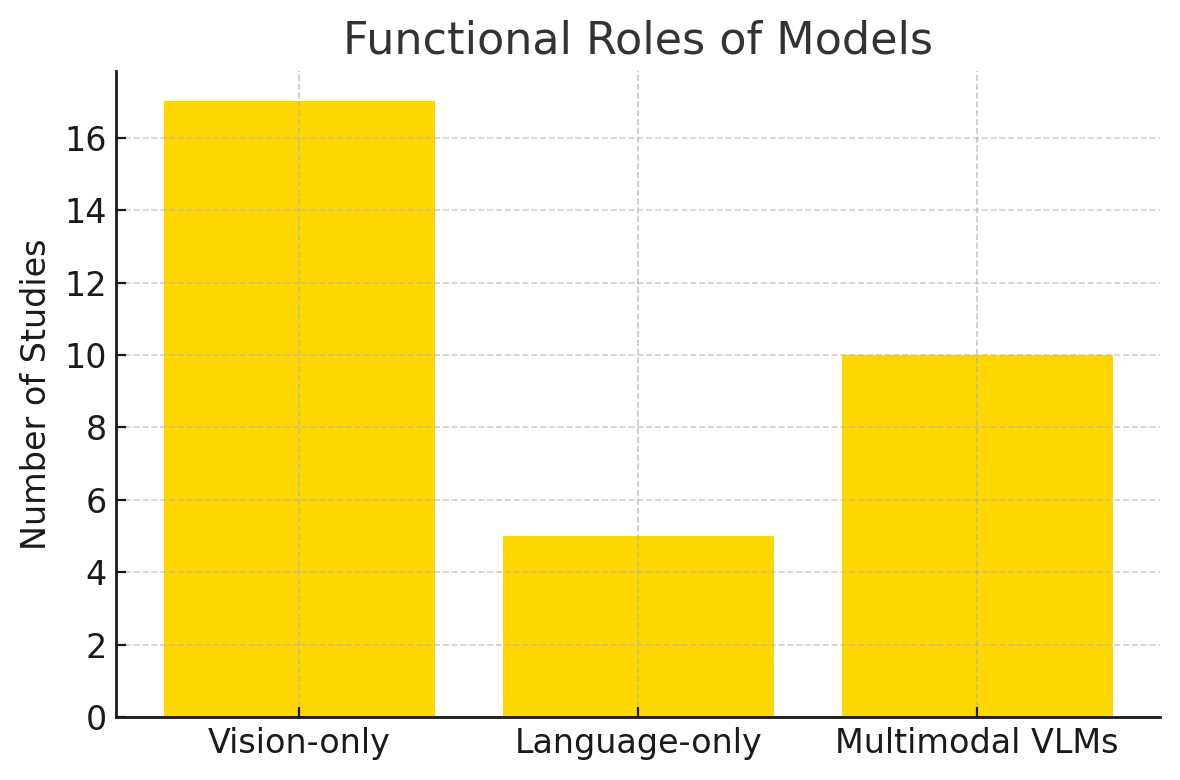}
    \caption{Functional roles of models across reviewed studies. Vision-only architectures are dominant, but VLMs and hybrid integrations are gaining traction.}
    \label{fig:model_roles_chart}
\end{figure}

\textbf{Usage Patterns and Gaps.} Across the corpus, vision-only approaches remain the most prevalent, reflecting their computational efficiency and availability of pretrained backbones. However, multimodal models are gaining traction in response to complex tasks such as geo-localization and navigation. Figure~\ref{fig:model_roles_chart} summarizes the distribution of functional model roles across studies.

\begin{figure}[h!]
    \centering
    \includegraphics[width=0.9\linewidth]{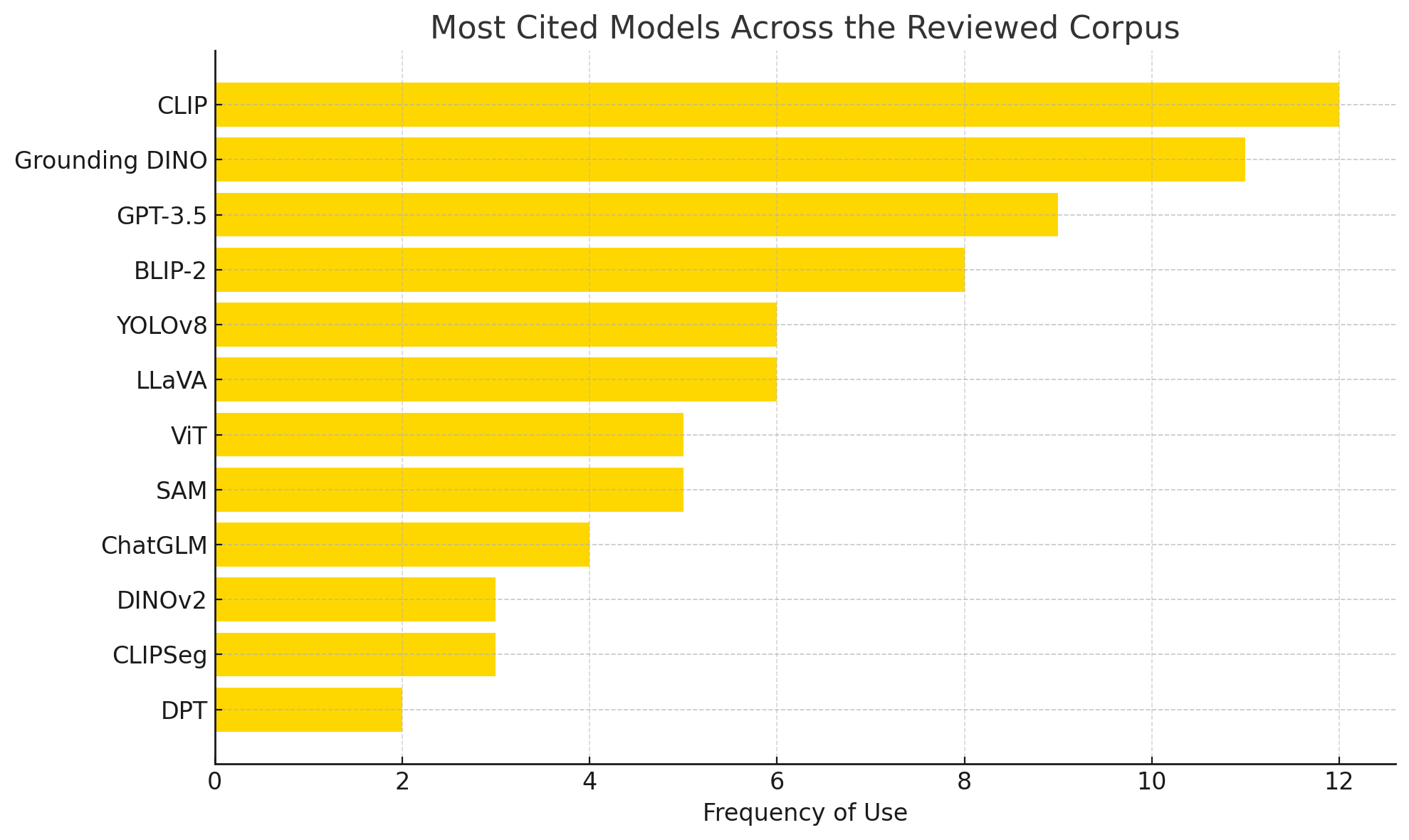}
    \caption{Most cited models across the reviewed corpus. CLIP, Grounding DINO, and GPT-3.5 are among the most prevalent.}
    \label{fig:specific_model_mentions}
\end{figure}

In terms of specific architectures, Figure~\ref{fig:specific_model_mentions} shows that CLIP, Grounding DINO, and GPT-3.5 are the most frequently used models. This trend reflects a strong preference for modular, general-purpose backbones that support zero-shot or prompt-based customization—but also reveals a reliance on a small subset of tools, potentially limiting innovation.

\section{Discussion}\label{sec4}

This systematic review of 32 recent papers reveals a fast-evolving landscape of zero-shot and few-shot vision-language models (VLMs) applied to urban computing. We categorized the literature into seven key domains: (1) Object Detection and Segmentation, (2) Urban Planning and Land Use Classification, (3) Navigation and Wayfinding, (4) Traffic Analysis and Transportation, (5) Scene Understanding and Perception, (6) Geo-localization and Location Finding, and (7) Urban Surveillance and Safety. Below, we synthesize findings within and across these categories, critically assess methodological limitations, and outline essential future directions to advance research and application.

\subsection{Performance Across Domains and Methodological Patterns}
The reviewed literature demonstrates both the promise and current limitations of applying zero-/few- shot and VLMs to urban tasks.

\textbf{Object Detection and Segmentation.} Seven studies addressed urban visual perception through detection and segmentation. Architectures combining vision-language features, such as SAM and Grounding DINO\cite{p2}, achieved strong IoU scores across common object classes. Notably, DHS-FSOD\cite{p5} demonstrated competitive mAP on industrial waste datasets. However, fisheye distortions, nighttime settings \cite{psingle}, and underrepresented categories (e.g., fire hydrants, crosswalks) significantly impaired performance\cite{p6}. Furthermore, reliance on synthetic datasets like CARLA\cite{p4} raises concerns about real-world generalization, especially in informal or heterogeneous urban morphologies not captured by MS COCO or Cityscapes.

\textbf{Urban Planning and Land Use Classification.} Studies in this category commonly used CLIP variants with structured prompts to classify street-level imagery into functional or perceptual urban zones. UrbanCLIP\cite{p9} and walkability prediction models\cite{p8} achieved high interpretability through prompt engineering. Yet, most experiments were confined to single-region datasets (e.g., Shenzhen, Amsterdam), lacked multilingual prompt adaptation, and showed reduced accuracy under facade occlusion or ambiguous mixed-use labeling\cite{p11}.

\textbf{Navigation and Wayfinding.} This group featured the most sophisticated multimodal pipelines, often integrating trajectory planning, language grounding, and visual input. PM-VLN\cite{p14} and ViPlanner\cite{p16} outperformed baselines on Touchdown and Matterport3D, yet evaluations were conducted on static, idealized conditions. The absence of ablation studies of dynamic elements, such as occlusion, signage variability, and human flow, limits the transferability to the real world. Moreover, models heavily depend on predefined camera positions and scene priors, which hampers scalability.

\textbf{Traffic Analysis and Transportation.} These studies utilized captioning-based VLMs (e.g., BLIP-2, LLaVA) to describe and interpret traffic scenarios. Divide \& Conquer\cite{p18} improved captioning precision via rule-based refinements, but scalability remains questionable. Few papers benchmarked real-time performance, and none integrated multimodal cues such as audio or sensor fusion, missing opportunities for robust safety-critical reasoning.

\textbf{Scene Understanding and Perception.} Works such as CityLLaVA\cite{p21} and CBFA\cite{p22} tackled zero-shot scene parsing and text recognition, often using domain adaptation and self-supervised learning. While mIoU and context-aware accuracy improved under limited supervision, these gains diminished in cross-domain transfer tests (e.g., SYNTHIA to Cityscapes). CLIPTER\cite{p24} addressed context in multilingual text recognition, yet still suffered from visual occlusion and signage variance. Overall, current methods are often too computationally heavy for deployment at scale.

\textbf{Geo-localization and Location Finding.} IM2City\cite{p27} and GeoReasoner\cite{p26} achieved promising top-1 and top-5 accuracy using reasoning-enhanced vision-language embeddings. However, generalization remains an issue across continents or low-data cities. Place recognition fails when urban morphology or signage differs markedly from training domains. Moreover, failure cases were rarely analyzed, and evaluation metrics like Recall@K neglect downstream risk implications of incorrect predictions.

\textbf{Urban Surveillance and Safety.} YOLO-based helmet detection and person re-ID\cite{p30, p29} achieved high mAP and Rank-1 accuracy, but potential overfitting to camera positions and identity pools poses concerns. Few-shot methods showed limited transferability, and privacy-related risks of re-ID were not addressed. These studies must also grapple with the socio-ethical implications of deploying surveillance in sensitive urban environments.

\subsection{Quantitative Trends and Metric Synthesis}

Across reviewed papers, performance reporting varies widely, complicating direct comparison. Key metrics include mAP, IoU, F1-score, task completion rate, and Recall@K, with limited use of confidence intervals or variance reporting. A summary of the best-reported metrics across all domains is presented in Table~\ref{tab:summary-metrics}, highlighting top-performing models and areas of progress.

\textbf{Object Detection and Segmentation.} DHS-FSOD\cite{p5} led with 32.82\% mAP@50. Grounded-SAM\cite{p2} reached IoUs of 0.78 for vehicles and infrastructure. Yet, real-world robustness (e.g., distorted lenses in\cite{p6}) remains underexplored, and most studies lack error bars or baseline comparisons.

\textbf{Urban Planning and Land Use Classification.} F1-scores were primary, with UrbanCLIP\cite{p9} achieving 0.82 for residential zones. Prompt-based models showed strong alignment with ontological labels, but lacked validation against human-annotated baselines in multilingual or mixed-use environments.

\textbf{Navigation and Wayfinding.} ViPlanner\cite{p16} and PM-VLN\cite{p14} reached 38.02\% semantic loss reduction and 33.4\% task completion respectively. However, no study reported variance across different navigation routes, obstacle configurations, or path lengths.

\textbf{Traffic Analysis and Transportation.} Dynamic prompting in\cite{p18} increased the AIC S2 score to 32.89. Metrics like CIDEr and BLEU were used but often lacked ground-truth alignment. Video-based methods rarely quantified latency, making deployment feasibility uncertain.

\textbf{Scene Understanding and Perception.} CBFA\cite{p22} achieved 57.52\% mIoU under 5-shot adaptation. Self-supervised models like SemiMTL\cite{p23} improved segmentation without labels but at high computational cost. Text-based methods reported weighted average error reductions, but rarely on cross-lingual benchmarks.

\textbf{Geo-localization and Location Finding.} IM2City\cite{p27} attained 85.9\% top-1 accuracy via linear probing, while GeoReasoner\cite{p26} achieved 75.2\% city-level accuracy. However, limited region diversity and opaque error analyses restrict generalizability claims.

\textbf{Urban Surveillance and Safety.} YOLOv8-based helmet detection\cite{p30} reported 92.3\% mAP@.05; person re-ID reached 95.9\% Rank-1\cite{p29}. Yet, studies overlooked evaluation under time drift, camera changes, and adversarial input.

\textbf{Synthesis.} Reporting remains inconsistent. Many studies omit baseline comparisons, confidence intervals, and detailed error analysis. A unified benchmark protocol with standardized metrics, failure case evaluation, and reproducibility checks would significantly advance the field (see Table~\ref{tab:summary-metrics} for consolidated results).

The metric landscape discussed above raises a natural next question: How are these models trained, validated, and deployed across datasets? To further contextualize the reported performance, we turn our attention to the patterns of dataset usage across the reviewed literature.

\begin{table}[h!]
\caption{Summary of Best-Reported Results Across Domains}
\label{tab:summary-metrics}
\centering
\begin{tabular}{p{0.20\textwidth}p{0.20\textwidth}p{0.15\textwidth}p{0.15\textwidth}p{0.1\textwidth}}
\toprule
\textbf{Domain} & \textbf{Best Method} & \textbf{Dataset} & \textbf{Metric(s)} & \textbf{Result} \\
\midrule
Object Detection & DHS-FSOD \cite{p5} & Huawei Waste, COCO & mAP@50 & 32.82\% \\
\midrule
Urban Planning & UrbanCLIP \cite{p9} & Baidu SVIs & F1 (residential) & 0.82 \\
\midrule
Navigation & PM-VLN \cite{p14} & Touchdown & Task Completion & 33.4\% \\
\midrule
Traffic Analysis & Divide \& Conquer \cite{p18} & WTS & AIC S2 Score & 32.89 \\
\midrule
Scene Understanding & CBFA \cite{p22} & SYNTHIA & mIoU & 57.52\% \\
\midrule
Geo-localization & IM2City \cite{p27} & Place Pulse 2.0 & Top-1 Acc. & 85.9\% \\
\midrule
Surveillance & Helmet Detection (YOLOv8) \cite{p30} & AIC Track 5 & mAP@.05 & 92.3\% \\
\bottomrule
\end{tabular}
\end{table}

\subsection{Dataset Usage Patterns}\label{sec4.2}

The dataset ecosystem underpinning urban zero-shot and few-shot learning is expanding in scope, yet structural shortcomings continue to limit model generalization, reproducibility, and real-world applicability. Based on our analysis of 32 studies, we identify recurring patterns that point to both promising directions and systemic blind spots in dataset design and evaluation protocols.

\textbf{Street-Level Dominance with Structural Bias.}
Datasets like Google Street View (GSV), Mapillary Vistas, and Baidu SVIs dominate vision-language applications due to their visual richness and alignment with human-centric urban scenes. However, these datasets exhibit pronounced access limitations (e.g., API restrictions, regional unavailability) and annotation inconsistencies. Consequently, while they serve as flexible platforms for spatial reasoning and pedestrian-level perception \cite{p10, p13, p16}, they often require manual curation and lack standardized benchmarking formats, raising concerns about replicability and bias propagation.

\textbf{Overreliance on Benchmark Datasets.}
Cityscapes and BDD100K remain the most frequently reused datasets for object detection, segmentation, and traffic scene understanding \cite{p5, p21,p24,p25}. Their consistent labeling supports reproducibility, but their geographic concentration in high-income Western contexts constrains cross-city generalization. The majority of reviewed studies perform training and evaluation within these narrow domains, risking overfitting to culturally and infrastructurally homogeneous settings. Models validated in Berlin or San Francisco, for instance, may struggle in cities with different signage, vehicle types, or pedestrian behavior patterns.

\textbf{Synthetic Data: Scale Without Grounded Realism.}
Synthetic datasets such as CARLA, SYNTHIA, and GTA5 are invaluable for simulating rare or hazardous conditions \cite{p8,p18,p24,p25}. They enable training in fog, night driving, or collision scenarios, settings where real data is difficult to obtain ethically or safely. Yet, synthetic-to-real domain gaps remain substantial. Most studies require extensive fine-tuning on real-world data post-training, limiting zero-shot applicability despite initial flexibility.

\textbf{Proprietary Data and Fragmented Evaluation.}
Several studies deploy internal datasets for niche use cases, e.g., waste detection \cite{p6}, fisheye distortion correction \cite{p17}, or helmet compliance \cite{p30}, that reflect real-world deployment challenges. However, the non-public nature of these datasets precludes benchmarking and community validation. This fragmentation indicates a broader lack of inclusive, task-aligned datasets that support deployment beyond the lab.

\textbf{Underdeveloped Evaluation for Cross-City and Temporal Generalization.}
A core finding is the near-total absence of structured evaluation for generalization across geographic, cultural, or infrastructural contexts. Although some studies attempt domain transfer (e.g., GTA5 to Cityscapes), few test models trained in one city against data from another with divergent features. Similarly, despite the availability of video datasets like BDD100K, temporal robustness is rarely addressed; most evaluations remain frame-by-frame, ignoring drift, consistency, or behavior modeling over time.

\textbf{Neglected Deployment Feasibility and Ethical Constraints.}
Surprisingly few works evaluate models in terms of runtime, hardware compatibility, sensor distortion, or annotation burden. Ethical dimensions, such as demographic representation, annotation fairness, privacy preservation (e.g., in facial data), and dataset provenance, are often overlooked or relegated to brief footnotes. This omission is particularly problematic for systems designed for deployment in public urban spaces, where regulatory and social implications are non-trivial.

\textbf{Remaining Gaps and Future Opportunities.} Despite improvements, current datasets lack temporal context, multilingual signage, and people-in-context annotations needed to simulate complex real-world interactions. The field would benefit from a unified, open-access benchmark combining multimodal inputs (image, video, text, GPS, audio) across globally diverse cities. Such a benchmark could enable rigorous, cross-region evaluation of generalization in zero-shot settings and support the development of globally inclusive urban AI systems.

\subsection{Limitations and Future Work}\label{sec4.2}

Despite rapid progress, the current landscape of vision-language models (VLMs) for urban environments is shaped by deep structural limitations, ranging from modality constraints to deployment fragility and ethical oversights. These challenges are not merely implementation hurdles but signal unresolved tensions in how the field conceptualizes generalization, context, and real-world applicability.

\textbf{Modality Gaps and Missing Context.}  
The overwhelming focus on static image-text pairs severely constrains the representational capacity of most urban VLM pipelines. Despite the ubiquity of real-world sensors that capture temporal sequences, depth maps, geolocation, and even ambient sound, these modalities are often excluded. This omission is especially problematic in domains such as autonomous navigation, traffic forecasting, crowd flow modeling, and multimodal safety analysis, where event progression, trajectory prediction, and ambient cues are critical.

Temporal dynamics, for example, are essential for understanding traffic congestion or pedestrian movement. Similarly, GPS data can ground visual inputs in spatial context, enabling fine-grained reasoning for applications like augmented reality wayfinding or real-time hazard detection. In edge deployments, such as AR glasses, mobile robots, or body-worn cameras, the under utilization of these rich, co-located modalities leads to poor performance and shallow reasoning.

Encouragingly, recent works such as OpenCity3D \cite{opencity3d} illustrates how 3D urban geometry can be integrated for depth-aware reasoning in planning and reconstruction, showing how modalities beyond RGB can enrich spatial inference. Similarly, although remote sensing datasets (e.g., for vegetation classification or infrastructure mapping) have traditionally relied on satellite or aerial imagery, advances in monocular depth estimation as Deep Anything V2 \cite{depth_anything_v2} allow RGB videos to approximate volumetric cues from the ground level, enabling scalable, street-view 3D understanding.

Yet despite these opportunities, urban VLM pipelines remain predominantly anchored in static image-text pairs. This leaves rich temporal and multimodal signals underutilized. While recent surveys and benchmarks in \textit{video foundation models (ViFMs)} \cite{videosurvey} have begun to explore how models can extract and reason over video sequences—including frame transitions, motion semantics, and multimodal streams—urban-specific models were left behind. These ViFMs fall into three classes: (1) image-based foundation models adapted for video, (2) video-native encoders, and (3) universal multimodal systems (UFMs) that jointly learn from image, video, text, and sometimes audio. Notably, the survey reveals that image-based foundation models still outperform native video encoders on many tasks, while UFMs exhibit superior performance across video benchmarks, especially when aligned with structured prompts and multimodal metadata.

However, such models remain largely absent from urban-specific implementations. This disconnect may, in part, reflect the inclusion criteria of this review, which focused on papers explicitly referencing vision-language and zero-/few-shot paradigms. Consequently, emerging multimodal architectures that do not fall under standard VLM nomenclature may be underrepresented. Future reviews might expand their scope to more fully capture advances in sensor fusion, temporal grounding, and edge-deployable urban reasoning.

Looking ahead, urban AI systems must move beyond single-frame inputs and embrace temporally grounded, sensor-fused pipelines. Robust city-scale reasoning will require synchronized input streams capable of modeling not only what is seen, but also when, where, and in what sequence. This opens new research directions in multimodal prompt conditioning, vision–IMU–LiDAR fusion, and multi-sensor attention alignment for dynamic urban environments.

\textbf{Metric Fragmentation and Inadequate Evaluation.}  
A consistent challenge across the reviewed literature is the lack of standardized evaluation protocols. While most studies report traditional metrics such as IoU, mAP, Recall@K, and F1-score, their implementation varies widely across domains and even within tasks—resulting in fragmented performance claims. For instance, segmentation studies often differ in threshold settings for IoU, and few report confidence intervals or per-class breakdowns, making comparisons across papers difficult if not misleading. In geo-localization and navigation tasks, some works rely on retrieval accuracy at different cutoffs (e.g., Recall@1, Recall@5), yet omit statistical variance or failure case analysis.

More concerning is the near-absence of \textit{deployment-centric metrics}. Only a handful of studies benchmark inference latency, energy consumption, or memory footprint—factors critical for real-world adoption in mobile or embedded systems. Even fewer assess robustness to adversarial examples, occlusions, or temporal drift—despite their importance in safety-critical urban settings like traffic analysis, crowd monitoring, or AR guidance.

Additionally, few works perform ablation studies or evaluate prompt sensitivity, despite prompt engineering being a central element of VLM design. Interpretability, fairness, and generalization outside the training region are rarely quantified. Without such evaluations, claims of “generalization” remain largely speculative.

This fragmentation prevents meaningful synthesis and makes it difficult to determine whether reported performance gains stem from model design, dataset bias, or evaluation selection. To address this, future studies should commit to transparent metric reporting, ideally incorporating \textit{unified benchmarks} that include both task accuracy and system-level metrics such as latency, power consumption, and fairness indicators. Inspired by efforts like EvalAI \cite{evalai}, the field would benefit from a shared evaluation toolkit with reproducible scripts, fixed test splits, and a public leaderboard.

\textbf{Overreliance on Resource-Intensive Architectures.}  
State-of-the-art performance in vision-language models is often achieved through extremely large and computationally expensive architectures, including GPT-4o, BLIP-2, and InternVL2. While powerful, these models are ill-suited for real-time, mobile, or embedded deployment, resulting in a widening chasm between innovation and practical application. This research-deployment gap is particularly acute in urban settings, where systems must operate within strict constraints on latency, power, and data privacy.

To bridge this divide, future work must embrace lightweight strategies. Promising directions include model pruning, quantization, and the use of fog or edge computing. Despite their potential, these techniques are rarely adopted in the urban VLM literature.

Recent advancements in \textbf{small language models (SLMs)}, such as Phi-2 \cite{phi15}, Mistral-7B \cite{mistral7b}, and TinyLLaMA \cite{tinyllama}, demonstrate that compact models can achieve strong performance on core language tasks with drastically reduced computational overhead. When integrated with efficient vision modules like YOLO-World \cite{YOLOWorld} or MobileSAMv2 \cite{mobilesamv2}, SLMs offer a pathway to build full-stack, multimodal reasoning pipelines capable of running directly on edge hardware, including drones, smartphones, and AR/VR systems. Florence-2 \cite{florence2} exemplifies this trend, providing a unified, edge-deployable architecture that supports diverse downstream tasks across modalities.

Yet despite these technological opportunities, few urban AI systems have capitalized on the synergy between SLMs and modular vision backbones. This oversight represents a critical bottleneck to achieving scalable, privacy-aware, and context-adaptive deployments in real-world environments. A shift toward \textit{resource-aware urban VLM design}, grounded in deployment constraints and powered by efficient multimodal models, will be essential to close the gap between lab-scale prototypes and field-ready solutions.

\textbf{Emerging Benchmarks for Language-Guided Urban Intelligence.}  
A promising development is the emergence of CityBench \cite{CityBench} and UrBench \cite{urbench}, two recent benchmarks originally designed for LLM and VLM evaluation, but highly relevant for zero-/few-shot visual systems due to their structured prompts and diverse visual data.

\textit{CityBench} represents a significant advancement in benchmarking urban reasoning capabilities in both LLMs and VLMs. It integrates multiple data modalities—geospatial data from OpenStreetMap, urban imagery from Google Maps, and user activity traces from platforms like Foursquare—into a unified simulation framework called \textit{CitySim}. This environment enables the generation of diverse, realistic urban scenarios that closely mimic real-world spatial layouts, semantic contexts, and human mobility patterns.

\begin{figure}[h!]
    \centering
    \includegraphics[height=0.6\linewidth]{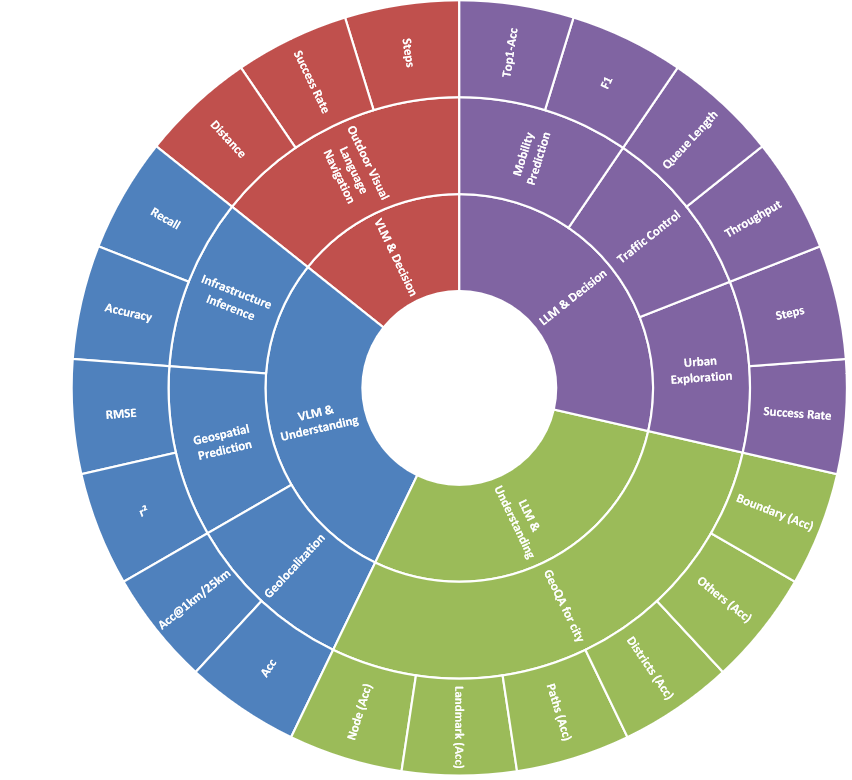}
    \caption{CityBench: Overview of the eight urban task families used to evaluate LLM/VLM capabilities across cities. Source from \cite{CityBench}}
    \label{fig:citybench_tasks}
\end{figure}

The benchmark is structured around eight distinct urban task groups, ranging from spatial layout understanding and POI (point-of-interest) recognition to urban planning decisions and real-time navigation (Figure~\ref{fig:citybench_tasks}). By embedding visual cues, natural language prompts, and geospatial metadata into each task, CityBench assesses how well a model can reason about urban dynamics in a multimodal and zero-shot setting.

Although originally designed to evaluate large foundation models, its simulation-based architecture, rich scene diversity, and prompt-grounded interactions make it highly suitable for evaluating visual models under constrained or domain-shifted conditions. For zero-shot and few-shot learning paradigms, CityBench provides a flexible and challenging testbed that captures the nuances of city-scale reasoning—enabling robust evaluation across city types, cultural norms, and environmental variables.

\begin{figure}[h!]
    \centering
    \includegraphics[width=1\linewidth]{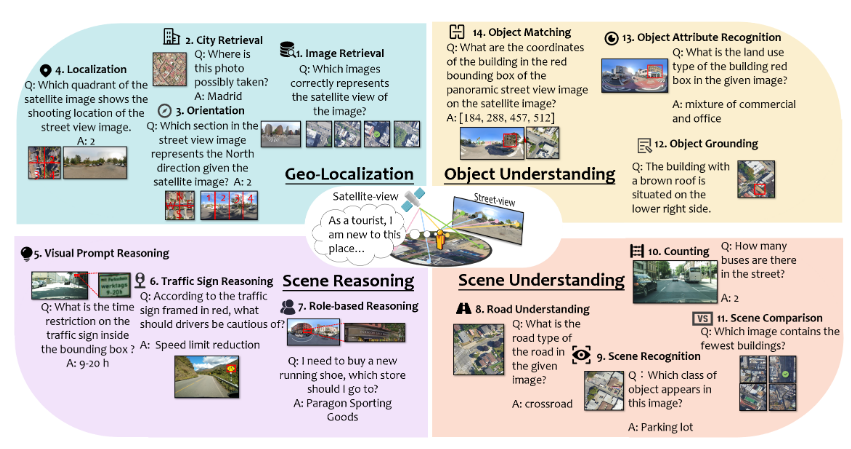}
    \caption{UrBench: Illustration of diverse vision-language tasks supported across 11 cities using street, satellite, and panoramic views. Source from \cite{urbench}}
    \label{fig:urbench_examples}
\end{figure}

\textit{UrBench} complements this with a QA-style dataset containing 11,600 samples across 14 vision-language tasks, covering multi-view urban imagery and tasks such as geo-localization, object grounding, and scene reasoning (Figure~\ref{fig:urbench_examples}). Designed to test spatial and semantic consistency across viewpoints, it exposes weaknesses in current multimodal systems—particularly when applied to cross-city generalization. 

Although originally developed to assess reasoning in LLMs, UrBench incorporates a wide range of visual content and structured prompts, making it highly valuable for zero-shot and few-shot evaluation. Its data is drawn from two sources: in-house collections (2,604 street-view and 4,239 satellite-view images from Google platforms, with geo-paired annotations) and curated public datasets including Cityscapes \cite{d1}, Mapillary Traffic Sign Dataset \cite{mapillaryTS}, VIGOR \cite{vigor}, and IM2GPS \cite{im2gps}. By leveraging spatial and temporal alignment, and integrating OpenStreetMap metadata, UrBench offers a benchmark rich in multimodal diversity and grounded urban semantics.

Together, these benchmarks signal a broader shift toward holistic, multimodal evaluation in urban AI. While their origin lies in LLM testing, their image-anchored tasks, diverse data modalities, and structured urban environments make them powerful resources for evaluating zero-/few-shot models.
Notably, both papers provide comprehensive benchmarking results on a wide range of contemporary models. CityBench evaluates the performance of LLMs such as Qwen2 series \cite{qwen2} and DeepSeekV2 \cite{DeepSeekV2AS}, as well as vision-language systems like LLaVA-NeXT-8B \cite{llavanext} and InternVL2 series \cite{internvl2}, across 8 distinct urban task types. Similarly, UrBench benchmarks 21 state-of-the-art multimodal models—including LLaVA series \cite{llavanext}, Gemini-1.5-Flash \cite{gemini}, GPT-4o \cite{gpt4}, and InstructBLIP \cite{instructblip}—across 14 spatially grounded visual tasks. These baseline scores serve as a valuable reference for future research, enabling reproducibility and comparative analysis of new approaches in urban zero-shot reasoning.

\textbf{Limited Experimentation with Edge Hardware.}  
Despite the availability of powerful mobile inference platforms, such as Google Coral TPU, NVIDIA Jetson, and Apple’s Neural Engine, few studies conduct simulation or deployment on such hardware. This lack of experimentation hampers our understanding of latency, thermal performance, and power trade-offs in real-world settings. Similarly, fog computing remains an underutilized paradigm in urban AI, despite its promise for low-latency, privacy-preserving inference at the edge. To bridge the research-deployment gap, future work must adopt end-to-end system designs explicitly grounded in deployment realities—considering not only model accuracy but also thermal constraints, communication overhead, and on-device privacy requirements.

\textbf{Ethical Blind Spots and Legal Oversights.}  
Despite the growing integration of AI into urban environments, ethical considerations remain an afterthought in the majority of vision-language model (VLM) research. Critical dimensions, such as algorithmic fairness, informed consent, and dataset provenance, are rarely embedded into model development pipelines. This disconnect raises serious concerns about public accountability and the socio-political legitimacy of urban AI deployments.

Notably, transparency around prompt engineering is frequently absent, and culturally sensitive benchmarks are rarely used. Intersectional bias audits and fairness evaluations, particularly relevant in public safety or infrastructure monitoring tasks, are often omitted, even when models are deployed in socially heterogeneous and surveillance-sensitive contexts.

The risks of such oversight are not merely theoretical. In New York City, the use of facial recognition in subway stations, implemented without public consultation, triggered investigations into civil liberties about racial bias and creep in surveillance \cite{newyork}. Similarly, the Metropolitan Police’s real-time facial recognition trials in London led to legal challenges and widespread protests, exposing a lack of transparency and insufficient governance mechanisms \cite{london}.

These documented incidents underscore the need for a proactive ethics-by-design approach. Emerging methods such as federated learning, differential privacy, and on-device inference offer technically feasible avenues to minimize centralized data collection risks. However, these strategies remain underexplored within the current urban VLM literature.

To close this ethical gap, future research should integrate:
\begin{itemize}
    \item Culturally diverse and demographically representative evaluation datasets,

    \item Standardized prompt documentation, including datasheets and "prompt cards" detailing assumptions and contexts,

    \item  Deployment-aware fairness protocols, sensitive to the urban environment’s socio-political dynamics.
\end{itemize}
Only by embedding ethics at each stage—from dataset curation to model inference—can we ensure that urban AI systems are not only technically capable, but socially aligned and publicly trustworthy.

\textbf{Toward a New Research Agenda.}  
To advance the field from a fragmented and prototype-heavy state to a robust, deployable, and ethically sound discipline, we propose a multidimensional research agenda centered on five foundational pillars:

\begin{itemize}
    \item \textit{SLM-VLM Hybrid Architectures.} Combine small language models (SLMs) with modular visual encoders and decoders to enable efficient multimodal reasoning on constrained hardware. Such combinations offer real-time, low-latency inference pipelines suitable for deployment on smartphones, drones, and wearable AR/VR devices.
    
    \item \textit{Unified Urban Benchmarks.} Develop evaluation suites that integrate multilingual prompts, multimodal sensor streams (e.g., image, video, audio, LiDAR), and culturally diverse geographic data. Standardizing such benchmarks will ensure reproducibility, foster cross-domain comparability, and promote robust generalization.

    \item \textit{Deployment-Centric Model Design.} Embed deployment constraints into the model development cycle, including hardware limitations, latency requirements, thermal budgets, and privacy considerations. This includes strategies like federated learning, quantization, pruning, and fog/edge inference pipelines that are particularly vital for public infrastructure and citizen-facing systems.

    \item \textit{Embedded Ethics and Compliance.} Integrate cultural robustness checks, algorithmic fairness evaluations, dataset consent tracking, and bias auditing into the core benchmarking and evaluation lifecycle. Ethical AI should not be a post-processing constraint but a co-evolving standard embedded into datasets, models, and deployment decisions.

    \item \textit{A Reproducible, Open Ecosystem.} Foster a culture of transparency through versioned datasets, dockerized baselines, public leaderboards, and shared evaluation code. Community-driven initiatives, such as urban-specific EvalAI tracks or CityBench extensions, can play a vital role in democratizing access and ensuring accountability.
\end{itemize}

In short, the field stands at a critical juncture. Expanding model capacity alone is no longer sufficient. To fulfill the promise of urban AI, future work must prioritize systems that are not only powerful, but also deployable, inclusive, interpretable, and ethically grounded.

\section{Conclusion}\label{sec13}

Vision-language models hold immense potential for transforming urban AI—from zero-shot detection in traffic footage to multimodal planning in city-scale simulations. However, our review reveals a field still shaped by prototype-centric designs, overreliance on large pretrained models, and limited attention to deployment and ethics. Current systems often fail to meet the constraints of urban infrastructures, neglecting edge deployment, cultural inclusivity, and real-time responsiveness.

To address these gaps, we propose a five-pillar research agenda centered on hybrid architectures, unified benchmarks, deployment-first design, embedded ethics, and reproducibility. The integration of lightweight SLMs with modular VLM pipelines, tested under culturally diverse and real-world conditions, offers a promising path forward.

As cities grow more dynamic and sensor-rich, the future of urban AI will depend on models that are not only intelligent, but also context-sensitive, explainable, and accessible. Building this future requires bridging technical excellence with social responsibility, a challenge that VLM research must now face.

\backmatter

\bmhead{Supplementary information}

If your article has accompanying supplementary file/s please state so here. 

Authors reporting data from electrophoretic gels and blots should supply the full unprocessed scans for key as part of their Supplementary information. This may be requested by the editorial team/s if it is missing.

Please refer to Journal-level guidance for any specific requirements.

\bmhead{Acknowledgements}

Acknowledgements are not compulsory. Where included they should be brief. Grant or contribution numbers may be acknowledged.

Please refer to Journal-level guidance for any specific requirements.

\section*{Declarations}

Some journals require declarations to be submitted in a standardised format. Please check the Instructions for Authors of the journal to which you are submitting to see if you need to complete this section. If yes, your manuscript must contain the following sections under the heading `Declarations':

\begin{itemize}
\item Funding
\item Conflict of interest/Competing interests (check journal-specific guidelines for which heading to use)
\item Ethics approval and consent to participate
\item Consent for publication
\item Data availability 
\item Materials availability
\item Code availability 
\item Author contribution
\end{itemize}

\noindent
If any of the sections are not relevant to your manuscript, please include the heading and write `Not applicable' for that section. 

\bigskip
\begin{flushleft}%
Editorial Policies for:

\bigskip\noindent
Springer journals and proceedings: \url{https://www.springer.com/gp/editorial-policies}

\bigskip\noindent
Nature Portfolio journals: \url{https://www.nature.com/nature-research/editorial-policies}

\bigskip\noindent
\textit{Scientific Reports}: \url{https://www.nature.com/srep/journal-policies/editorial-policies}

\bigskip\noindent
BMC journals: \url{https://www.biomedcentral.com/getpublished/editorial-policies}
\end{flushleft}

\begin{appendices}

\section{Dataset Inventory}\label{secA1}

This appendix provides a comprehensive tabulation of the datasets referenced across the 32 studies reviewed in this paper. The datasets are grouped by their primary data modality and use case, including:

\begin{itemize}
    \item Street-level imagery;
    \item Aerial and top-down views;
    \item Simulated urban environments;
    \item Proprietary/internal datasets.
\end{itemize}

Each table includes dataset characteristics, usage examples, and associated references. While the main text highlights the most frequently adopted or methodologically influential datasets, this appendix serves as a reference resource to support replication, dataset comparison, and further exploration by the research community.

\begin{table}
\caption{Datasets in Real-World Street-Level Imagery.}\label{tab:dataset-real}
\centering
\begin{tabular}{p{0.15\textwidth}p{0.15\textwidth}p{0.35\textwidth}p{0.1\textwidth}p{0.1\textwidth}}
\toprule
\textbf{Dataset} & \textbf{Primary Use} & \textbf{Main Characteristics} & \textbf{Used In}  \\
\midrule
Cityscapes & Semantic Segmentation, Driving & 5,000 finely annotated street-view images from 50 cities & \cite{p24, p25}  \\
\midrule
BDD100K & Object Detection, Driving Analytics & 100K driving videos under varied conditions (day/night, weather) & \cite{p21, p23} \\
\midrule
Google Street View & Geo-localization, Urban Planning & Panoramic city-scale imagery & \cite{p2,p11,p15,p28} \\
\midrule
Mapillary Vistas & Urban Planning, Segmentation & Crowdsourced urban imagery with segmentation labels & \cite{p10} \\
\midrule
Baidu SVIs & Geo-localization & 226K street-level images from Shenzhen & \cite{p11}  \\
\midrule
ActivityNet Captions & Video Captioning & 20K annotated videos for captioning tasks & \cite{p1} \\
\midrule
Charades-STA & Temporal Video Understanding & Grounded temporal actions in video & \cite{p1}  \\
\midrule
Touchdown & Navigation, Language Grounding & 6,525 natural-language routes in NYC & \cite{p16}  \\
\midrule
TR-NY-PIT-Central & Navigation & 17,000 routes with image-language pairs & \cite{p16} \\
\midrule
MC-10 & Landmark Detection & 8,100 landmark images with annotations & \cite{p16}  \\
\midrule
Place Pulse 2.0 & Perception, Geo-localization & 111K images with perceptual ratings & \cite{p13}, \cite{p26} \\
\midrule
GTSRB & Traffic Sign Recognition & 43 traffic sign categories from German roads &  \cite{p21}  \\
\midrule
Market-1501 & Person Re-ID & 32K person re-ID images, 6 cameras & \cite{p29}  \\
\midrule
DukeMTMC-reID & Person Re-ID & 36K re-ID images across 8 cameras & \cite{p29} \\
\midrule
CUHK (x3) & Person Re-ID & Multi-domain person re-ID datasets & \cite{p29} \\
\bottomrule
\end{tabular}
\end{table}

\begin{table}
\caption{Datasets in Aerial or Top-Down Urban Views.}\label{tab:dataset-top}
\centering
\begin{tabular}{p{0.15\textwidth}p{0.15\textwidth}p{0.35\textwidth}p{0.1\textwidth}p{0.1\textwidth}}
\toprule
\textbf{Dataset} & \textbf{Primary Use} & \textbf{Main Characteristics} & \textbf{Used In}\\
\midrule
Potsdam & Depth Estimation, Segmentation & High-res orthophotos with semantic and depth labels & \cite{p25}  \\
\midrule
Vaihingen & Segmentation, Cross-domain Learning & Urban imagery for cross-domain learning & \cite{p25}  \\
\bottomrule
\end{tabular}
\end{table}

\begin{table}
\caption{Datasets in Simulated Urban Environments.}\label{tab:dataset-simulated}
\centering
\begin{tabular}{p{0.15\textwidth}p{0.15\textwidth}p{0.35\textwidth}p{0.1\textwidth}p{0.1\textwidth}}
\toprule
\textbf{Dataset} & \textbf{Primary Use} & \textbf{Main Characteristics} & \textbf{Used In}  \\
\midrule
CARLA & Driving Simulation, Domain Adaptation & Synthetic driving scenes with rare events & \cite{p8,p18}  \\
\midrule
NVIDIA Omniverse & Navigation Simulation & Modular 3D environments for planning/navigation & \cite{p18}  \\
\midrule
Warehouse Simulations & Navigation Planning & Custom 3D indoor environments & [18]  \\
\midrule
GTA5 & Domain Adaptation, Segmentation & 24K synthetic scenes for segmentation & \cite{p24}  \\
\midrule
SYNTHIA & Domain Adaptation, Segmentation & Synthetic urban segmentation dataset & \cite{p24,p25}  \\
\midrule
Synscapes & Segmentation, Semi-supervised Learning & Synthetic RGB-D scenes with rich metadata  \\
\midrule
Matterport3D & Indoor Navigation & Indoor 3D reconstructions for navigation tasks & \cite{p18}  \\
\bottomrule
\end{tabular}
\end{table}

\begin{table}
\caption{Proprietary or Internal Datasets}\label{tab:dataset-private}
\centering
\begin{tabular}{p{0.15\textwidth}p{0.15\textwidth}p{0.35\textwidth}p{0.1\textwidth}p{0.1\textwidth}}
\toprule
\textbf{Dataset} & \textbf{Primary Use} & \textbf{Main Characteristics} & \textbf{Used In} \\
\midrule
FishEye8K & Distortion Modeling, Object Detection & 8K distorted images, 157K annotations & \cite{p7} \\
\midrule
Huawei Waste Dataset & Waste Detection & 29 annotated waste object categories & \cite{p6} \\
\midrule
Woven Traffic Safety Dataset & Traffic Captioning & Internal/external videos with rich labels & \cite{p19,p20,p23}  \\
\midrule
AI City Challenge 2023 & Violation Detection & 200 helmet detection videos (Track 5) & \cite{p30} \\
\bottomrule
\end{tabular}
\end{table}





\end{appendices}


\bibliography{sn-bibliography}

\end{document}